\documentclass[10pt,journal]{IEEEtran}

\usepackage{amsmath,amssymb}
\usepackage[pdftex]{graphicx}
\usepackage{hyperref}
\usepackage{algorithm} 
\usepackage{algorithmicx}
\usepackage{algpseudocode}
\usepackage{multirow}
\usepackage{cite}
\usepackage{url}
\usepackage{enumerate}
\usepackage{alltt}

\usepackage{color}

\usepackage{amsmath,amsthm,amssymb}
\usepackage{bbm}
\usepackage{mathtools}

\newtheorem{theorem}{Theorem}
\newtheorem{lemma}{Lemma}

\newtheorem{assumption}{Assumption}

\newcommand{\eg}{\emph{e.g.}}
\newcommand{\ie}{\emph{i.e.}}
\newcommand{\aka}{\emph{a.k.a.}}
\newcommand{\etal}{\emph{et al.}}
\newcommand{\vct}[1]{\ensuremath{\boldsymbol{#1}}}
\newcommand{\mat}[1]{\ensuremath{\mathtt{#1}}}
\newcommand{\set}[1]{\ensuremath{\mathcal{#1}}}
\newcommand{\con}[1]{\ensuremath{\mathsf{#1}}}
\newcommand{\T}{\ensuremath{^\top}}

\newcommand{\erf}{\text{erf}}
\newcommand{\argmin}{\operatornamewithlimits{\arg\,\min}}
\newcommand{\bmat}[1]{\begin{bmatrix}#1\end{bmatrix}}

\begin{document}

\title{Randomized Prediction Games for \\ Adversarial Machine Learning}

\author{Samuel~Rota~Bul\`o,~\IEEEmembership{Member,~IEEE,}
Battista~Biggio,~\IEEEmembership{Member,~IEEE,}
Ignazio~Pillai,~\IEEEmembership{Member,~IEEE,}
Marcello~Pelillo,~\IEEEmembership{Fellow,~IEEE}
Fabio~Roli,~\IEEEmembership{Fellow,~IEEE}
\thanks{S. Rota Bul\`o is with ICT-Tev, Fondazione Bruno Kessler, Trento, Italy}
	\thanks{M. Pelillo is with DAIS, Universit\`a Ca' Foscari, Venezia, Italy}
	\thanks{B. Biggio, I. Pillai, F. Roli are with DIEE, University of Cagliari, Italy}}

\IEEEcompsoctitleabstractindextext{
\begin{abstract}
In spam and malware detection, attackers exploit randomization to obfuscate malicious data and increase their chances of evading detection at test time; e.g., malware code is typically obfuscated using random strings or byte sequences to hide known exploits. Interestingly, randomization has also been proposed to improve security of learning algorithms against evasion attacks, as it results in hiding information about the classifier to the attacker. Recent work has proposed game-theoretical formulations to learn secure classifiers, by simulating different evasion attacks and modifying the classification function accordingly. However, both the classification function and the simulated data manipulations have been modeled in a deterministic manner, without accounting for any form of randomization. In this work, we overcome this limitation by proposing a randomized prediction game, namely, a non-cooperative game-theoretic formulation in which the classifier and the attacker make randomized strategy selections according to some probability distribution defined over the respective strategy set. We show that our approach allows one to improve the trade-off between attack detection and false alarms with respect to state-of-the-art secure classifiers, even against attacks that are different from those hypothesized during design, on application examples including handwritten digit recognition, spam and malware detection.
\end{abstract}

}

\maketitle

\IEEEdisplaynotcompsoctitleabstractindextext

\section{Introduction} \label{sect:introduction}

Machine-learning algorithms have been increasingly adopted in \emph{adversarial settings} like spam, malware and intrusion detection.
However, such algorithms are not designed to operate against intelligent and adaptive attackers, thus making them inherently vulnerable to carefully-crafted attacks.
Evaluating security of machine learning against such attacks and devising suitable countermeasures, are two among the main open issues under investigation in the field of \emph{adversarial machine learning} \cite{dalvi04,lowd05,barreno06-asiaccs,cardenas-ws06,laskov09,huang11,biggio14-tkde,biggio14-svm-chapter,biggio13-ecml,bruckner09,bruckner12}. 
In this work we focus on the issue of designing secure classification algorithms against \emph{evasion attacks}, \ie{}, attacks in which malicious samples are manipulated at test time to evade detection. This is a typical setting, \eg{}, in spam filtering, where spammers manipulate the content of spam emails to get them past the anti-spam filters \cite{wittel04,lowd05,dalvi04,kolcz09,biggio10-ijmlc}, or in malware detection, where hackers obfuscate \emph{mal}icious soft\emph{ware} (\emph{malware}, for short) to evade detection of either known or zero-day exploits \cite{chris05,fogla06,biggio14-svm-chapter,biggio13-ecml}. 
Although out of the scope of this work, it is worth mentioning here another pertinent attack scenario, referred to as \emph{classifier poisoning}. Under this setting, the attacker can manipulate the training data to mislead classifier learning and cause a denial of service; \eg, by increasing the number of misclassified samples~\cite{nelson08,rubinstein09,huang11,biggio14-tkde,biggio12-icml,biggio15-icml}.

To date, several authors have addressed the problem of designing secure learning algorithms to mitigate the impact of evasion attacks~\cite{dalvi04,globerson06-icml,teo08,dekel10,huang11,bruckner09,bruckner11,bruckner12,grosshans13,vamvoudakis14,zhang15} (see Sect.~\ref{sec:related-work} for further details).
The underlying rationale of such approaches is to learn a classification function that accounts for potential malicious data manipulations at test time. To this end, the interactions between the classifier and the attacker are modeled as a game in which the attacker manipulates data to evade detection, while the classification function is modified to classify them correctly. This essentially amounts to incorporating knowledge of the attack strategy into the learning algorithm. 
However, both the classification function and the simulated data manipulations have been modeled in a deterministic manner, without accounting for any form of \emph{randomization}. 

Randomization is often used by attackers to increase their chances of evading detection, \eg{}, malware code is typically obfuscated using random strings or byte sequences to hide known exploits, and spam often contains bogus text randomly taken from English dictionaries to reduce the ``spamminess'' of the overall message.
Surprisingly, randomization has also been proposed to improve classifier security against evasion attacks~\cite{biggio08-spr,barreno06-asiaccs,huang11}.
In particular, it has been shown that \emph{randomizing} the learning algorithm may effectively hide information about the classification function to the attacker, requiring her to select a less effective attack (manipulation) strategy. In practice, the fact that the adversary may not know the classification function exactly (\ie, in a deterministic sense) decreases her (expected) payoff on each attack sample. This means that, to achieve the same expected evasion rate attained in the deterministic case, the attacker has to increase the number of modifications made to the attack samples~\cite{biggio08-spr}.

Motivated by the aforementioned facts, in this work we 
generalize \emph{static prediction games}, \ie, the game-theoretical formulation proposed by Br\"uckner \etal~\cite{bruckner09,bruckner12}, to account for randomized classifiers and data manipulation strategies.
For this reason, we refer to our game as a \emph{randomized prediction game}. A randomize prediction game is a non-cooperative game between a \emph{randomized} learner and a \emph{randomized} attacker (also called data generator), where the player's strategies are replaced with probability distributions defined over the respective strategy sets.
Our goal is twofold. We do not only aim to assess whether randomization helps achieving a better trade-off in terms of false alarms and attack detection (with respect to state-of-the-art \emph{secure} classifiers), but also whether our approach remains more secure against attacks that are different from those hypothesized during design.
In fact, given that our game considers randomized players, it is reasonable to expect that it may be more robust to potential deviations from its original hypotheses about the players' strategies.

The paper is structured as follows. Randomized prediction games are presented in Sect.~\ref{sec:rand-pred-games}, where sufficient conditions for the existence and uniqueness of a Nash equilibrium in these games are also given.
In Sect.~\ref{sec:case_study} we focus on a specific game instance involving a linear Support Vector Machine (SVM) learner, for which we provide an effective method to find an equilibrium by overcoming some computational problems. 
We discuss how to enable the use of nonlinear (kernelized) SVMs in Sect.~\ref{sec:kernelization}.
In Sect.~\ref{sec:discussion} we report a simple example to intuitively explain how the proposed methods enforce security in adversarial settings. 
Related work is discussed in Sect.~\ref{sec:related-work}.
In  Sect.~\ref{sec:exp} we empirically validate the soundness of the proposed approach on an handwritten digit recognition task, and on realistic adversarial application examples involving spam filtering and malware detection in PDF files. Notably, to evaluate robustness of our approach and state-of-the-art secure classification algorithms, we also consider attacks that deviate from the models hypothesized during classifier design.
Finally, in Sect.~\ref{sec:conclusions}, we summarize our contributions and sketch potential directions for future work.


\section{Randomized Prediction Games}
\label{sec:rand-pred-games}

Consider an adversarial learning setting involving two actors: a \emph{data generator} and a \emph{learner}.\footnote{We adopt here the same terminology used in~\cite{bruckner12}.} The data generator produces at \emph{training time} a set
$\hat{\set D}=\{\hat{\vct x_i},y_i\}_{i=1}^{\con n}\subseteq \set X\times\set Y$ of $\con n$ training samples, sampled from an unknown probability distribution.
Sets $\set X$ and $\set Y$ denote respectively the input and output spaces of the learning task.
At \emph{test time}, the data generator modifies the samples in $\hat{\set D}$ to form a new dataset $ {\set D}\subseteq\set X\times\set Y$, reflecting a test distribution, which differs in general from the training distribution and it is not available at training time. We assume binary learners, \ie{} $\set Y=\{-1,+1\}$, and we assume also that the data transformation process leaves the labels of the samples in $\hat{\set D}$ unchanged, \ie{}, ${\set D}=\{({\vct x_i},y_i)\}_{i=1}^{\con n}$. Hence, a perturbed dataset will simply be represented in terms of a tuple ${\mat X}=({\vct x}_1,\dots,{\vct x}_{\con n})\in\set X^\con n$, each element being the perturbation of the original input sample $\hat{\vct x_i}$, while we implicitly assume the label to remain $y_i$.
The role of the learner is to classify samples $\vct x\in\set X$ according to the prediction function
$h(\vct x)=\text{sign}[f(\vct x;\vct w)]$, which is expressed in terms of a linear generalized decision function $f(\vct x;\vct w)=\vct w\T\phi(\vct x)$, where $\vct w\in\mathbb R^{\con m}$, $\vct x\in\set X$ and $\phi:\set X\rightarrow\mathbb R^{\con m}$ is a \emph{feature map}. 

Static prediction games have been introduced in \cite{bruckner12} by modeling the learner and the data generator as players of a non-cooperative game that we identify as $l$-player and $d$-player, respectively.
The strategies of $l$-player correspond to the parametrizations $\vct w$ of the prediction function $f$.
The strategies of the data generator, instead, are assumed to live directly in the feature space, by regarding $\dot{\mat X}=(\dot{\vct x}_1\T,\dots,\dot{\vct x}_\con n\T)\T\in\mathbb R^{\con m\con n}$ as a data generator strategy, where $\dot{\vct x}_i=\phi(\vct x_i)$.
By doing so, the decision function $f$ becomes linear in either players' strategies. 
Each player is characterized also by a \emph{cost} function that depends on the strategies played by either players. 
The cost function of $d$-player and $l$-player are denoted by $c_d$ and $c_l$, respectively, and are given by
\begin{equation}
  c_l(\vct w,\dot{\mat X})=\rho_l\Omega_l(\vct w)+\sum_{i=1}^{\con n} \ell_l(\vct w\T\dot{\vct x}_i,y_i) \,,
	\label{eq:costs_l}\\
\end{equation}
\begin{equation}
	c_d(\vct w,\dot{\mat X})=\rho_d\Omega_d(\dot{\mat X})+\sum_{i=1}^{\con n} \ell_d(\vct w\T\dot{\vct x}_i,y_i)\,,
	\label{eq:costs_d}
\end{equation}
where $\vct w\in\mathbb R^\con m$ is the strategy of $l$-player, $\dot{\mat X}\in\mathbb R^{\con m\con n}$ is the strategy of the $d$-player, and $y_i$ denotes the label of $\dot{\vct x}_i$ as per $\hat{\set D}$.
Moreover, $\rho_{d/l}> 0$ is a trade-off parameter, $\ell_{d/l}(\vct w\T\dot{\vct x}_i,y)$ measures the loss incurred by the $l$/$d$-player when the decision function yields $\vct w\T\dot{\vct x_i}$ for the $i$th training sample while the true label is $y$, and $\Omega_{d/l}$ can be regarded as a penalization for playing a specific strategy. For the $d$-player, this term quantifies the cost of perturbing $\hat{\set D}$ in feature space.

The goal of our work is to introduce a randomization component in the model of \cite{bruckner12}, particularly to what concerns the players' behavior.
To this end, we take one abstraction step with respect to the aforementioned prediction game, where we let the learner and the data generator sample their playing strategy in the prediction game
from a parametrized distribution, under the assumption that they are expected cost-minimizing (\aka{} expected utility-maximizing). By doing so, we introduce a new non-cooperative game that we call \emph{randomized prediction game} between the $l$-player 
and the $d$-player with strategies being mapped to the possible parametrizations of the players' respective distributions, and cost functions being expected costs under the same distributions. 

\vspace{-10pt}
\subsection{Definition of randomized prediction game}
Consider a prediction game as described before.
We inject randomness in the game by assigning each player a parametrized probability distribution, \ie, $p_l(\vct w;\vct \theta_l)$ for the learner and $p_d(\dot{\mat X};\vct \theta_d)$ for the data generator, that governs the players' strategy selection.
Players are allowed to select the parametrization $\vct\theta_l$ and $\vct\theta_d$ for the respective distributions. 
For any choice of $\vct\theta_l$, the $l$-player plays a strategy $\vct w$ sampled from $p_l(\cdot;\vct\theta_l)$. 
Similarly, for any choice of $\vct\theta_d$, the $d$-player plays a strategy $\dot{\mat X}$ sampled from $p_d(\cdot;\vct\theta_d)$.
If the players adhere to the new rules, we obtain a \emph{randomized prediction game}.

A \emph{randomized prediction game} is a non-cooperative game between a learner ($l$-player) and data generator ($d$-player) that has the following components:
\begin{enumerate}
  \item an underlying prediction game with cost functions $c_{l/d}(\vct w,\dot{\mat X})$ as defined in \eqref{eq:costs_l} and \eqref{eq:costs_d},
	\item two parametrized probability distributions $p_{l/d}(\cdot;\vct\theta_{l/d})$ with parameters in $\Theta_{l/d}$,
  \item $\Theta_{l/d}$ are non-empty, compact and convex subsets of a finite-dimensional metric space $\mathbb R^{\con s_{l/d}}$.
\end{enumerate}
The sets of parameters $\Theta_{l/d}$ are the \emph{pure strategy} sets (\aka{} action spaces) for the $l$-player and $d$-player, respectively.
The costs functions of the two players, which quantify the cost that each player incurs when a strategy profile $(\vct\theta_l,\vct\theta_d)\in\Theta_l\times\Theta_d$ is played, coincide with 
 the expected costs, denoted by $\overline c_{l/d}(\vct\theta_l,\vct\theta_d)$, 
  that the two players have in the underlying prediction game if strategies are sampled from $p_l(\cdot;\vct\theta_l)$ and $p_d(\cdot;\vct\theta_d)$, according to the expected cost-minimizing hypothesis:
\begin{align}
	\overline c_l(\vct\theta_l,\vct\theta_d)&=\mathbb E_{\substack{\vct w\sim p_l(\cdot;\vct\theta_l)\\\dot{\mat X}\sim p_d(\cdot;\vct\theta_d)}}[c_l(\vct w,\dot{\mat X})]\,,
	\label{eq:costs_mixed_l}\\
	\overline c_d(\vct\theta_l,\vct\theta_d)&=\mathbb E_{\substack{\vct w\sim p_l(\cdot;\vct\theta_l)\\ \dot{\mat X}\sim p_d(\cdot;\vct\theta_d)}}[c_d(\vct w,\dot{\mat X})]\,,
	\label{eq:costs_mixed_d}
\end{align}
where $\mathbb E[\cdot]$ denotes the expectation operator.
We assume $\overline c_{l/d}$ to be well-defined functions, \ie{} the expectations to be finite for any $(\vct\theta_l,\vct\theta_d)\in\Theta_l\times\Theta_d$.
To avoid confusion between $c_{l/d}$ and $\overline c_{l/d}$, in the remainder of this paper we will refer them respectively as cost functions, and \emph{expected} cost functions.

By adhering to a non-cooperative setting, the two players involved in the prediction game are not allowed to communicate and they play their strategies 
simultaneously. Each player has complete information of the game setting by knowing the expected cost function and strategy set of either 
players. Under rationality assumption, each player's interest is to achieve the greatest personal advantage, \ie{}, to incur the lowest possible cost. Accordingly, 
the players are prone to play a \emph{Nash equilibrium}, which
in the context of our randomized prediction game is a strategy profile $(\vct\theta^\star_l,\vct\theta^\star_d)\in\Theta_l\times\Theta_d$ such that no player is interested in changing his/her 
own playing strategy. In formal terms, this yields:
\begin{equation}
\vct\theta^\star_l\in\argmin_{\vct\theta_l\in\Theta_l}\overline c_l(\vct\theta_l,\vct\theta^\star_d)\,,\quad\vct\theta^\star_d\in\argmin_{\vct\theta_d\in\Theta_d}\overline c_d(\vct\theta^\star_l,\vct\theta_d)\,.
	\label{eq:nash}
\end{equation}

\vspace{-10pt}
\subsection{Existence of a Nash equilibrium}
The existence of a Nash equilibrium of a randomized prediction game is not granted in general.
A sufficient condition for the existence of a Nash equilibrium is thus given below.

\begin{theorem}[Existence]
  A randomized prediction game admits at least one Nash equilibrium if 
  \begin{enumerate}[(i)]
    \item $\overline c_{l/d}$ are continuous in $\Theta_l\times\Theta_d$,
    \item $\overline c_l(\cdot,\vct\theta_d)$ is quasi-convex in $\Theta_l$ for any $\vct\theta_d\in\Theta_d$, 
    \item $\overline c_d(\vct\theta_l,\cdot)$ is quasi-convex in $\Theta_d$ for any $\vct\theta_l\in\Theta_l$.
  \end{enumerate}
	\label{prop:cond}
\end{theorem}
\begin{IEEEproof}
  The result follows directly from the Debreu-Glicksberg-Fan Theorem \cite{glicksberg52}.
\end{IEEEproof}

\vspace{-10pt}
\subsection{Uniqueness of a Nash equilibrium}
In addition to the existence of a Nash equilibrium, it is of interest to investigate if the equilibrium is unique.
However, determining tight conditions that guarantee the uniqueness of the Nash equilibrium for any randomized prediction game is challenging; in particular, due to the additional dependence on a probability distribution for the learner and the data generator. 

We will make use of a classical result due to Rosen \cite{Ros65} to formulate sufficient conditions for the uniqueness of the Nash equilibrium of randomized prediction games in terms of 
the so-called pseudo-gradient of the game, defined as 
\begin{equation}
\overline{g}_{\vct r} = \begin{bmatrix}
r_{l} \nabla_{\vct \theta_{l}} \bar c_{l} \\
r_{d} \nabla_{\vct \theta_{d}} \bar c_{d}
 \end{bmatrix} \enspace , 
 \label{eq:gr}
\end{equation}
with any fixed vector $\vct r=[r_{l}, r_d]\T\geq \vct 0$.
Specifically, a randomized prediction game admits a unique Nash equilibrium if the following assumption is verified
\begin{assumption}
  \leavevmode
  \begin{enumerate}[(i)]
    \item $\overline c_{l/d}$ are twice differentiable in $\Theta_l\times\Theta_d$,
    \item $\overline c_{l}(\cdot,\vct\theta_d)$ is convex in $\Theta_l$ for any $\vct\theta_d\in\Theta_d$, 
    \item $\overline c_{d}(\vct\theta_l,\cdot)$ is convex in $\Theta_{d}$ for any $\vct\theta_l\in\Theta_l$,
  \end{enumerate}
  \label{ass1}
\end{assumption}
\noindent and $\overline{g}_{\vct r}$ is strictly monotone for some fixed $\vct r>\vct 0$, \ie, 
\[
  [\overline{g}_{\vct r} (\vct \theta_l,\vct \theta_d)-\overline{g}_{\vct r} (\vct \theta'_l,\vct \theta'_d)]\T\bmat{\vct\theta_l-\vct\theta'_l\\\vct\theta_d-\vct\theta'_d}>0 \, ,
\]
for any distinct strategy profiles $(\vct\theta_l,\vct\theta_d),(\vct\theta'_l,\vct\theta'_d)\in\Theta_l\times\Theta_d$.\footnote{Assumption 1.(i) could be relaxed to continuously differentiable.}

In his paper, Rosen provides also a useful sufficient condition that guarantees a strictly monotone pseudo-gradient. This requires the Jacobian of the pseudo-gradient, \aka{} \emph{pseudo-Jacobian}, given by
\begin{equation}
  \overline J_{\vct r}=\bmat{r_l\nabla^2_{\vct\theta_l\vct \theta_l}\overline c_l&r_l\nabla^2_{\vct\theta_l\vct \theta_d}\overline c_l\\
  r_d\nabla^2_{\vct\theta_d\vct \theta_l}\overline c_d&r_d\nabla^2_{\vct\theta_d\vct \theta_d}\overline c_d}\,,
  \label{eq:Jr}
\end{equation}
  to be positive definite.
\begin{theorem}
A randomized prediction game admits a unique Nash equilibrium if Assumption \ref{ass1} holds, 
and the pseudo-Jacobian $\overline J_{\vct r}(\vct\theta_l,\vct\theta_d)$ is positive definite for all $(\vct\theta_l,\vct\theta_d)\in\Theta_l\times\Theta_d$ and some fixed $\vct r>\vct 0$.
  \label{thm:unique_jr}
\end{theorem}
\begin{IEEEproof}
  Under Assumption~\ref{ass1},
  the positive definiteness of $\overline J_{\vct r}$
  for all strategy profiles and some fixed vector $\vct r>\vct 0$ implies the strict monotonicity of $\overline g_{\vct r}$, which in turn implies the uniqueness of the Nash equilibrium \cite[Thm.~6]{Ros65}.
\end{IEEEproof}

In the rest of the section, we provide sufficient conditions that ensure the positive definiteness of the pseudo-Jacobian and thus the uniqueness of the Nash equilibrium via Thm.~\ref{thm:unique_jr}.
To this end we decompose $\bar c_{l/d}(\vct\theta_l,\vct\theta_d)$ as follows
\begin{equation}
\begin{aligned}
    \bar c_l(\vct\theta_l,\vct\theta_d)&=\rho_l\overline\Omega_l(\vct\theta_l)+\bar L_l(\vct\theta_l,\vct\theta_d)\,,\\
    \bar c_d(\vct\theta_l,\vct\theta_d)&=\rho_d\overline\Omega_d(\vct\theta_d)+\bar L_d(\vct\theta_l,\vct\theta_d)\,,
\end{aligned}
\label{eq:c_scomp}
\end{equation}
where $\overline\Omega_{l/d}$ and $\bar L_{l/d}$ are the expected regularization and loss terms given by
\begin{align*}
\overline\Omega_l(\vct\theta_l)&=\mathbb E_{\vct w\sim p_l(\cdot,\vct\theta_l)}[\Omega_l(\vct w)]\,,\\
\overline\Omega_d(\vct\theta_d)&=\mathbb E_{\dot{\mat X}\sim p_d(\cdot,\vct\theta_d)}[\Omega_d(\dot{\mat X})]\,,\\
\bar L_{l}(\vct\theta_l,\vct\theta_d)&=\mathbb E_{\substack{\vct w\sim p_l(\cdot;\vct\theta_l)\\\dot{\mat X}\sim p_d(\cdot;\vct\theta_d)}}
\left[\sum_{i=1}^\con n\ell_{l}(\vct w\T\vct x_i,y_i)\right]\,,\\
\bar L_d(\vct\theta_l,\vct\theta_d)&=\mathbb E_{\substack{\vct w\sim p_l(\cdot;\vct\theta_l)\\\dot{\mat X}\sim p_d(\cdot;\vct\theta_d)}}
\left[\sum_{i=1}^\con n\ell_d(\vct w\T\vct x_i,y_i)\right]\,.
\end{align*}
Moreover, we require the following convexity and differentiability conditions on $\overline\Omega_{l/s}$ and $\bar L_{l/d}$:
\begin{assumption}\label{assx}\hfill
    \begin{enumerate}[(i)]
        \item $\overline \Omega_{l/d}$ is strongly convex and twice continuously differentiable in $\Theta_{l/d}$,
        \item $\bar L_l(\cdot,\vct\theta_d)$ is convex and twice continuously differentiable in $\Theta_l$ for all $\vct\theta_d\in\Theta_d$, and
        \item $\bar L_d(\vct\theta_l,\cdot)$ is convex and twice continuously differentiable in $\Theta_d$ for all $\vct\theta_l\in\Theta_l$.
    \end{enumerate}
\end{assumption}
Finally, we introduce some quantities that are used in the subsequent lemma, which gives sufficient conditions for the positive-definiteness of the pseudo-Jacobian:
\begin{align*}
    \lambda^\Omega_{l}&=\inf_{\vct\theta_{l}\in\Theta_l}\lambda_{\text{min}}\left[\nabla^2_{\vct\theta_l\vct\theta_l}\overline\Omega_l(\vct\theta_l)\right]\,,\\
\lambda^\Omega_d&=\inf_{\vct\theta_d\in\Theta_d}\lambda_{\text{min}}\left[\nabla^2_{\vct\theta_d\vct\theta_d}\overline\Omega_d(\vct\theta_d)\right]\,,\\
\lambda^L_l&=\inf_{(\vct\theta_l,\vct\theta_d)\in\Theta_l\times\Theta_d}\lambda_{\text{min}}\left[\nabla^2_{\vct\theta_l\vct\theta_l}\overline L_l(\vct\theta_l,\vct\theta_d)\right]\,,\\
\lambda^L_d&=\inf_{(\vct\theta_l,\vct\theta_d)\in\Theta_l\times\Theta_d}\lambda_{\text{min}}\left[\nabla^2_{\vct\theta_d\vct\theta_d}\overline L_d(\vct\theta_l,\vct\theta_d)\right]\,,\\
\tau&=\sup_{(\vct\theta_l,\vct\theta_d)\in\Theta_l\times\Theta_d}\lambda_{\text{max}}\left[R(\vct\theta_l,\vct\theta_d)R(\vct\theta_l,\vct\theta_d)\T\right]\,,
\end{align*}
where 
$R(\vct\theta_l,\vct\theta_d)=\frac{1}{2}\left[\nabla^2_{\vct\theta_l\vct\theta_d}\bar L_l(\vct\theta_l,\vct\theta_d)\T+\nabla^2_{\vct\theta_d\vct\theta_l}\bar L_d(\vct\theta_l,\vct\theta_d)\right]$ and
$\lambda_{\text{max/min}}$ give the maximum/minimum eigenvalue of the matrix in input.
Note that the quantities listed above are finite and positive if Assumption~\ref{assx} holds,
given the compactness of $\Theta_{l/d}$.
\begin{lemma}
    If Assumption~\ref{assx} holds and 
    \[
(\rho_l\lambda^\Omega_l+\lambda^L_l)(\rho_d\lambda^\Omega_d+\lambda^L_d)>\tau
    \]
    then the pseudo-Jacobian $\overline{ J}_r(\vct\theta_l,\vct\theta_d)$ is positive definite for all $(\vct\theta_l,\vct\theta_d)\in\Theta_l\times\Theta_d$ by taking $\vct r=(1,1)\T$.
    \label{thm:new_cond}
\end{lemma}
\begin{IEEEproof}
    The pseudo-Jacobian in \eqref{eq:Jr} can be written as follows given the decomposition of $\bar c_{l/d}$ in \eqref{eq:c_scomp}:
    \[
        \mat J_r=\bmat{\rho_l\nabla_{\vct\theta_l\vct\theta_l}^2\overline\Omega_l+
                       \nabla_{\vct\theta_l\vct\theta_l}^2\bar L_l &\nabla_{\vct\theta_l\vct\theta_d}^2\bar L_l\\
                       \nabla_{\vct\theta_d\vct\theta_l}^2\bar L_d&
                       \rho_d\nabla_{\vct\theta_d\vct\theta_d}^2\overline\Omega_d+\nabla_{\vct\theta_d\vct\theta_d}^2\bar L_d 
    }\,,
    \]
    where we omitted the arguments of $\overline\Omega_{l/d}$ and $\bar L_{l/d}$ for notational convenience.
    Let us denote by $\mat J_r^{ll}$, $\mat J_r^{ld}$, $\mat J_r^{dl}$, and $\mat J_r^{dd}$ the four matrices composing $\mat J_r$ (in top-down, left-right order).

    Consider the following matrix:
    \[
        \mat H=\bmat{\mat H^{ll}&\mat H^{ld}\\\mat H^{dl}&\mat H^{dd}}=\bmat{
    \rho_l\lambda_l^\Omega+\lambda_l^L&R(\vct\theta_l,\vct\theta_d)\T\\R(\vct\theta_l,\vct\theta_d)&\rho_d\lambda_d^\Omega+\lambda_d^L
}.
    \]
    Then we have for all $\vct t=(\vct t_l\T,\vct t_d\T)\neq \vct 0$
    \begin{multline*}
        \vct t\T\mat J_r\vct t=\vct t\frac{\mat J_r+\mat J_r\T}{2}\vct t\T\\=\underbrace{\vct t_l\mat J_r^{ll}\vct t_l}_{\geq\vct t_l\mat H^{ll}\vct t_l} +\underbrace{\vct t_d\mat J_r^{dd}\vct t_d}_{\geq\vct t_d\T\mat H^{dd}\vct t_d}+\vct t_l\T\underbrace{\frac{\mat J_r^{ld}+\mat J_r^{dl\top}}{2}}_{\mat H^{ld}+\mat H^{dl\top}}\vct t_d\geq\vct t\T\mat H\vct t\,,
\end{multline*}
where the under-braced relations follow from the definitions of $\lambda_{l/d}^\Omega$, $\lambda_{l/d}^L$ and $R$.
    Accordingly, the positive-definiteness of $\mat J_r$ can be derived from the positive-definiteness of matrix $\mat H$. To prove the latter, we will show that all roots of the characteristic polynomial $\text{det}(\mat H-\lambda\mat I)$ of $\mat H$ are positive. By properties of the determinant\footnote{$\det\bmat{a \mat I&\mat B\T\\\mat B&d \mat I}=\det(a\mat I)\det(d\mat I-\frac{1}{a}\mat B\mat B\T)$ and if $\mat U\mat S\mat U\T$ is the eigendecomposition of $\mat B\mat B\T$ then the latter determinant becomes $\det(\mat U(d\mat I-\frac{1}{a}\mat S)\mat U\T)=\det(d\mat I-\frac{1}{a}\mat S)$  } we have
    \begin{multline*}
        \det(\mat H-\lambda\mat I)=\det( (\rho_l\lambda^\Omega_l+\lambda^L_l-\lambda)\mat I)\\
        \cdot\det\left((\rho_d\lambda^\Omega_d+\lambda^L_d-\lambda)\mat I-\frac{\mat S}{\rho_l\lambda^\Omega_l+\lambda^L_l-\lambda}\right)\,,
    \end{multline*}
    where $\mat S$ is a diagonal matrix with the eigenvalues of $R(\vct\theta_l,\vct\theta_d)R(\vct\theta_l,\vct\theta_d)\T$.
    The roots of the first determinant term are all equal to $\rho_l\lambda^\Omega_l+\lambda^L_l$, which is positive because $\rho_l>0$ by construction and $\lambda^\Omega_l>0$ follows from the strong-convexity of $\overline\Omega_l$ in Assumption~\ref{assx}-i.
    As for the second determinant term, take the $i$th diagonal element $\mat S_{ii}$ of $\mat S$. Then two roots are the solution of the following quadratic equation
    \[
        \lambda^2-\lambda(a+b)+ab-\mat S_{ii}=0\,,
    \]
    which are given by
    \[
        \lambda^{(i)}_{1,2}=a+b\pm\sqrt{(a-b)^2+4\mat S_{ii}}\,.
    \]
    where $a=\rho_l\lambda^\Omega_l+\lambda^L_l$ and $b=\rho_d\lambda^\Omega_d+\lambda^L_d$.
    Among the two, $\lambda^{(i)}_2$ (the one with the minus) is the smallest one, which is strictly positive if
    \[
        ab=(\rho_l\lambda^\Omega_l+\lambda^L_l)(\rho_d\lambda^\Omega_d+\lambda^L_d)>\mat S_{ii}\,.
    \]
    Since the condition has to hold for any choice of the eigenvalue $\mat S_{ii}$ in the right-hand-side of the inequality, we take the maximum one $\max_i\mat S_{ii}$, which coincides with $\lambda_\text{max}(R(\vct\theta_l,\vct\theta_d)R(\vct\theta_l,\vct\theta_d)\T)$. We further maximize the latter quantity with respect to $(\vct\theta_l,\vct\theta_d)\in\Theta_l\times\Theta_d$, because we want the result to hold for any parametrization. Therefrom we recover the variable $\tau$ and the condition
$(\rho_l\lambda^\Omega_l+\lambda^L_l)(\rho_d\lambda^\Omega_d+\lambda^L_d)>\tau$, which guarantees that all roots of the characteristic polynomial of $\mat H$ are strictly positive for any choice of $(\vct\theta_l,\vct\theta_d)\in\Theta_l\times\Theta_d$ and, hence, $\overline J_r$ is positive definite over $\Theta_l\times\Theta_d$.
\end{IEEEproof}

In addition to Lem.~\ref{thm:new_cond}, we provide in the supplementary material alternative (stronger) sufficient conditions, which generalize the ones given in \cite{bruckner12}.

\vspace{-10pt}
\subsection{Finding a Nash equilibrium} 

\begin{algorithm}[t]
\caption{Extragradient descent (adapted from~\cite{bruckner12})}
\label{alg:eg-nash}
\begin{algorithmic}[1]
\Require Cost functions $\bar c_{l/d}$; 
 parameter spaces $\Theta_{l},\Theta_{d}$; a small positive constant $\epsilon$.
\Ensure The optimal parameters $\vct \theta_{l}, \vct \theta_{d}$.
\State Randomly select $\vct \theta^{(0)} = (\vct\theta^{(0)}_{l}, \vct\theta^{(0)}_{d}) \in \Theta_{l} \times \Theta_{d}$.  
\State Set iteration count $k = 0$, and select $\sigma, \beta \in (0,1)$.
\State Set $\vct r=(r_{l},r_{d})^{\T} = (1, \rho_{l} / \rho_{d})^{\T}$.
\Repeat
	\State Set
	$ \vct d^{(k)} =
	\Pi_{\Theta_{l} \times \Theta_{d}} \left( \vct \theta^{(k)} - \overline{\vct g}_{\vct r}\left(\vct \theta^{(k)}_{l}, \vct \theta^{(k)}_{d} \right ) \right) - \vct \theta^{(k)}$.
	\State Find maximum step size $t^{(k)} \in \{ \beta^{p} | p \in \mathbb N \}$ s.t.
	\[
		- \overline{\vct g}_{\vct r} \left ( \bar{\vct \theta}^{(k)}_{l},
			\bar{\vct \theta}^{(k)}_{d} \right )^{\T}
			\vct d^{(k)} \geq
			\sigma \left( \left \| \vct d^{(k)} \right 
			\|^{2}_{2} \right),
	\] where
	$\bar{\vct \theta}^{(k)} = \vct \theta^{(k)} + t^{(k)}\vct d^{(k)}$.
	\State Set $\eta^{(k)} =  - \frac{t^{(k)}}{\left \| \overline{\vct g}_{\vct r} \left ( \bar{\vct \theta}^{(k)}_{l},
			\bar{\vct \theta}^{(k)}_{d} \right ) \right \|^{2}_{2}}
			\overline{\vct g}_{\vct r} \left ( \bar{\vct \theta}^{(k)}_{l},
			\bar{\vct \theta}^{(k)}_{d} \right )^{\T}
			\vct d^{(k)}$.

	\State Set
		${\vct \theta}^{(k+1)}=
		\Pi_{\Theta_{l} \times \Theta_{d}} \left( \vct \theta^{(k)}
		- \eta^{(k)} \overline{\vct g}_{\vct r} \left ( \bar{\vct \theta}^{(k)}_{l},
			\bar{\vct \theta}^{(k)}_{d} \right )
		\right)   
		$.
	\State Set $k = k+1$.
\Until{ $ 
		\left \|  {\vct \theta}^{(k)} - {\vct \theta}^{(k-1)}  \right \|^{2}_{2}  \leq \epsilon
	  $  }
\State \Return ${\vct \theta}_{l} = {\vct \theta}^{(k)}_{l}$, ${\vct \theta}_{d} = {\vct \theta}^{(k)}_{d}$
\end{algorithmic}
\end{algorithm}

From the computational perspective, we can find a Nash equilibrium in our game by exploiting algorithms similar to the ones adopted for static prediction games~\cite{bruckner12}.
In particular, we consider a modified extragradient descent algorithm~\cite{bruckner12,zhu93,HarPan90} that finds a solution to the following variational inequality problem, provided that $\overline{g}_{\vct r}$ is continuous and monotone: 
\begin{equation}
\overline{g}_{\vct r}(\vct\theta^{\star}_{l},\vct\theta^{\star}_{d})^{\T} \left( \vct \theta - \vct \theta^{\star} \right) \geq 0 \enspace , \forall (\vct \theta_l,\vct\theta_d) \in \Theta_{l} \times \Theta_{d} \, ,
\end{equation}
where $\vct\theta=[\vct\theta_l^{\top},\vct\theta_d^{\top}]\T$ and similarly for $\vct\theta^\star$.
Any solution $\vct \theta^{\star}$ to this problem can be shown to correspond bijectively to a Nash equilibrium of a game having $\overline{g}_{\vct r}$ as pseudo-gradient~\cite{HarPan90,bruckner12}.

If Theorem~\ref{prop:cond} holds, the pseudo-Jacobian $\bar J_{\vct r}$ can be shown to be positive semidefinite, and $\overline g_{\vct r}$ is thus continuous and monotone.
Hence, the variational inequality can be solved by the modified extragradient descent algorithm given as Algorithm~\ref{alg:eg-nash}, which is guaranteed to converge to a Nash equilibrium point~\cite{zhu93,GeiKan99}.
The algorithm generates a sequence of feasible points whose distance from the equilibrium solution is monotonically decreased.
It exploits a projection operator $\Pi_{\Theta_{l} \times \Theta_{d}}(\vct \theta)$ to map the input vector $\vct \theta$ onto the closest admissible point in $\Theta_{l} \times \Theta_{d}$,
and a simple line-search algorithm to find the maximum step $t$ on the descent direction $\vct d$.\footnote{We refer the reader to \cite{bruckner12,zhu93,HarPan90} (and references therein) for detailed proofs that derive conditions for which $\vct d$ is effectively a descent direction.}

In the next section, we apply our randomized prediction game to the case of linear SVM learners, and compute the corresponding pseudo-gradient, as required by Algorithm~\ref{alg:eg-nash}.

\section{Randomized Prediction Games for Support Vector Machines}
\label{sec:case_study}

In this section, we consider a randomized prediction game involving a linear SVM learner~\cite{CorVap95}, and Gaussian distributions as the underlying probabilities $p_{l/d}$. 

\textbf{The learner.} The decision function of the learner is of the type $f(\vct x;\vct w)=\vct w\T\phi(\vct x)$ where the feature map is given by $\phi(\vct x)=\bmat{\vct x\T& 1}\T$. For convenience, we consider a decomposition of $\vct w$ into $\bmat{\tilde{\vct w}\T&b}\T$, where $\tilde{\vct w}\in\mathbb R^{\con m-1}$ and $b\in\mathbb R$. Hence, the decision function can also be written as $f(\vct x;\vct w)=\tilde{\vct w}\T\vct x+b$. 
Accordingly, the input space $\set X$ is a $({\con m-1})$-dimensional vector space, \ie{} $\set X\subseteq\mathbb R^{\con m-1}$.
The distribution $p_l$ for the learner is assumed to be Gaussian. In order to guarantee the theoretical existence of the Nash equilibrium through Thm.~\ref{prop:cond}, we assume the parameters 
of the Gaussian distribution to be bounded. For the sake of clarity, we use in this section axis-aligned Gaussians (\ie{} with diagonal covariance matrices) for our analysis, 
even though general covariances could be adopted as well. 
Under these assumptions, we define the strategy set for the learner as $\Theta_l=\left\{ (\vct \mu_{\vct w},\vct \sigma_{\vct w})\in\mathbb R^{\con m}\times\mathbb R_+^{\con m}\right\}\cap \set B_l$,
where $\set B_l\subset\mathbb R^{\con m}\times\mathbb R_+^{\con m}$ is an application-dependent non-empty, convex, bounded set, restricting the set of feasible parameters.
The parameter vectors $\vct \mu_{\vct w}$ and $\vct\sigma_{\vct w}$ encode the mean and standard deviation of the axis-aligned Gaussian distributions.
The loss function $\ell_l$ of the learner corresponds to the hinge loss of the SVM, \ie{}, $\ell_l(z,y)=[1-z y]_+$ with $[z]_+=\max(0,z)$, while the strategy penalization term $\Omega_l(\vct w)$ is the squared Euclidean norm of $\tilde{\vct w}$.
As a result, the cost function $c_l$ corresponds to the C-SVM objective function, and it is convex in $\vct w$:
\begin{equation}
	c_l(\vct w,{\mat X})=\frac{\rho_l}{2}\Vert\tilde{\vct w}\Vert^2+ \sum_{i=1}^{\con n}[1-y_i(\tilde{\vct w}\T {\vct x}_i+b)]_+ \,.
	\label{eq:cl_SVM}
\end{equation}

\textbf{The data generator.} For convenience, we consider $\mat X$ rather than $\dot {\mat X}$ as the quantity undergoing the randomization.
This comes without loss of generality, because there is a one-to-one correspondence between $\dot{\vct x_i}$ and $\vct x_i$ if we consider the linear feature map $\dot{\vct x_i}=\phi(\vct x_i)=[\vct x_i\T,1]\T$.
Moreover, we assume that samples $\vct x_i$ can be perturbed independently.
Accordingly, the distribution $p_d$ for the data generator factorizes as
$p_d({\mat X};\vct \theta_d)=\prod_{i=1}^{\con n}p_d\left({\vct x_i}; \vct \theta^{(i)}_d\right)$,
where $\vct \theta_d=(\vct\theta^{(1)}_d,\dots,\vct \theta^{(\con n)}_d)$. 
We consider $p_d\left({\vct x_i}; \vct\theta^{(i)}_d\right)$ to be a $\con k$-variate axis-aligned Gaussian distribution with bounded mean and standard 
deviation given by $\vct\theta^{(i)}_d=(\vct \mu_{\vct x_i},\vct \sigma_{\vct x_i})$.
In summary, the strategy set adopted for the data generator is given by $\Theta_d=\prod_{i=1}^{\con n}\Theta_d^{(i)}$, where
$\Theta_d^{(i)}=\left\{ (\vct \mu_{\vct x_{i}},\vct \sigma_{\vct x_{i}})\in \mathbb R^{\con k}\times\mathbb R^{\con k}_+\right\}\cap\set B_d$.
Here, $\set B_d\subset\mathbb R^{\con k}\times\mathbb R^{\con k}_+$ is a non-empty, convex, bounded set.
The loss function $\ell_d$ of the data generator is the hinge loss under wrong labelling, \ie{}, $\ell_d(z,y)=[1+z y]_+$. 
In this way the data generator is penalized if the learner correctly classifies a sample point. Finally, the strategy penalization function $\Omega_d$ is the squared Euclidean distance of the perturbed samples in ${\mat X}$ from the ones in the original training set $\hat{\set D}$, 
\ie{} $\Omega_d({\mat X})=\sum_{i=1}^\con n\Vert \vct x_i-\hat{\vct x_i}\Vert^2$.
The resulting cost function $c_d$ is convex in ${\mat X}$:
\begin{multline}
    c_d(\vct w,{\mat X})=\frac{\rho_d}{2}\sum_{i=1}^\con n\Vert \vct x_i-\hat{\vct x_i}\Vert^2\\+\sum_{i=1}^{\con n}[1+y_i(\tilde{\vct w}\T {\vct x}_i+b)]_+\,.
	\label{eq:cd_SVM}
\end{multline}

\textbf{Existence of a Nash equilibrium.}
The proposed randomized prediction game for the SVM learner admits at least one Nash equilibrium. 
This can be proven by means of Thm.~\ref{prop:cond}. Indeed, the required continuity of $\bar c_{l/d}$ hold and,
as for the quasi-convexity conditions, we can rewrite \eqref{eq:costs_mixed_l} as follows by exploiting the fact that $p_l$ is a Gaussian distribution with mean $\vct \mu_{\vct w}$ and standard deviation $\vct\sigma_{\vct w}$:
\begin{equation}
	\overline c_l(\vct\theta_l,\vct\theta_d)=\mathbb E_{\substack{\vct z\sim\mathcal N(\vct 0,I)\\{\mat X}\sim p_d(\cdot;\vct\theta_d)}}[c_l(\vct \mu_{\vct w}+D(\vct \sigma_{\vct w})\vct z,{\mat X})]\,,
	\label{eq:cost_learner}
\end{equation}
where $\set N(\vct 0,I)$ is a $\con m$-dimensional standard normal distribution and $D(\vct \sigma_{\vct w})$ is a diagonal matrix having $\vct \sigma_{\vct w}$ on the diagonal. Since $c_l$ is convex in its first argument and convexity is preserved under addition of convex functions, positive rescaling, and composition with linear functions, we have that $\overline c_l$ is convex (and thus quasi-convex) in $\vct\theta_l=(\vct \mu_{\vct w},\vct\sigma_{\vct w})$. As for the quasi-convexity condition of the data generator's cost, we can exploit the separability of $c_d$ to rewrite \eqref{eq:costs_mixed_d} as follows:
\[
	\overline c_d(\vct\theta_l,\vct\theta_d)=\sum_{i=1}^{\con n}\mathbb E_{\substack{\vct w\sim p_l(\cdot;\vct\theta_l)\\\vct z\sim \mathcal N(\vct 0,I)}}[c_d^{(i)}(\vct w,\vct\mu_{\vct x_i}+D(\vct \sigma_{\vct x_i})\vct z)]\,,
\]
where
\[
c_d^{(i)}(\vct w,{\vct x})=\frac{\rho_d}{2}\Vert {\vct x}-\hat{\vct x_i}\Vert^2+ [1+y_i({\vct w}\T \tilde{\vct x}_i+b)]_+\,.
	\]
	Since $c_d^{(i)}$ is convex in its second argument, by following the same reasoning used to show the quasi-convexity of the learner's expected 
	cost, we have that each expectation in $\overline c_d$ is convex in $\vct\theta^{(i)}_d=(\vct \mu_{\vct x_i},\vct \sigma_{\vct x_i})$, $1\leq i\leq\con n$. As a consequence, $\overline c_d$ is convex and, hence, quasi-convex in $\vct\theta_d$, being the sum of convex functions.

\textbf{Uniqueness of a Nash equilibrium.}
In the previous section we have shown that $\bar c_{l}(\cdot,\vct\theta_d)$ and  $\bar c_{d}(\vct\theta_l,\cdot)$ are convex as required by Assumption~\ref{ass1}-(ii-iii). In particular we have that the single expected regularization terms $\overline\Omega_{l/d}(\cdot)$ and loss terms $\bar L_{l}(\cdot,\vct\theta_d)$, $\bar L_{d}(\vct\theta_l,\cdot)$ are convex as well.
Moreover, they are twice-continuously differentiable by having Gaussian distributions for $p_{l/d}$.
It is then sufficient to have $\overline\Omega_{l/d}$ are strongly convex to prove the uniqueness of the Nash equilibrium via Lem.~\ref{thm:new_cond}. While it is easy to see that $\overline\Omega_{d}$ is strongly convex, we have that
$\overline\Omega_{l}$ is \emph{not} strongly convex with respect to $b$ due to the presence of an \emph{unregularized} bias term $b$ in the learner.
The problem derives from the fact that the SVM itself may not have a unique solution when the bias term is present and non-regularized (see \cite{abe02,burges99} for a characterization of the degenerate cases).
As a result, the proposed game is not guaranteed to have a unique Nash equilibrium in its actual form.
On the other hand, a unique Nash equilibrium may be obtained by either considering an unbiased SVM, \ie, by setting $b=0$ as in \cite{bruckner12}, or a \emph{regularized} bias term, \eg, by adding $\frac{\varepsilon}{2} b^{2}$ to the learner's objective function with $\varepsilon>0$. In both cases, 
all conditions that ensure the uniqueness of the Nash equilibrium via Thm.~\ref{thm:unique_jr} and Lem.~\ref{thm:new_cond} would be satisfied, under proper choices of $\rho_{l/d}$.

It is worth noting however that the necessary and sufficient conditions under which a biased (non-regularized) SVM has no unique solution are quite restricted~\cite{abe02,burges99}. For this reason, we believe that uniqueness of the Nash equilibrium could be proven also for the biased SVM under mild assumptions. 
However, this requires considerable effort in trying to relax the sufficiency conditions of Rosen~\cite{Ros65}, which is beyond the scope of our work. We thus leave this challenge to future investigations.
Moreover, we believe that enforcing a unique Nash equilibrium in our game by making the original SVM formulation strictly convex may lead to worse results, similarly to exploiting convex approximations to solve originally non-convex problems in machine learning~\cite{collobert06,bengio07}.
For the above reasons, in this paper, we choose to retain the original SVM formulation for the learner, by sacrificing the uniqueness of the Nash Equilibrium. 
We nevertheless provide in Sect.~\ref{sec:discussion} a discussion of why having a unique Nash Equilibrium is not so important in practice for our game, and we empirically show in Sect.~\ref{sec:exp} that our approach can anyway achieve competitive performances with respect to other state-of-the-art approaches.

The rest of this section is devoted to showing how to compute the pseudo-gradient \eqref{eq:gr} by providing explicit formulae for $\nabla_{\vct\theta_l}\bar c_l$ and $\nabla_{\vct\theta_d}\bar c_d$.

\vspace{-10pt}
\subsection{Gradient of the learner's cost}
\label{ssec:learner}

In this section, we focus on computing the gradient $\nabla_{\vct\theta_l}\bar c_l(\vct\theta_l,\vct\theta_d)$,
where $\bar c_l$ is defined as in \eqref{eq:cl_SVM}.
By properties of expectation and since $\vct w$ follows an axis-aligned Gaussian distribution with mean $\vct \mu_{\vct w}$ and standard deviation $\vct \sigma_{\vct w}$, we can reduce the cost of the learner to:
\begin{multline}
  \bar c_l(\vct\theta_l,\vct\theta_d)=
\frac{\rho_l}{2}\left( \Vert\vct \mu_{\tilde{\vct w}}\Vert^2 + \Vert\vct\sigma_{\tilde{\vct w}} \Vert^2 \right) \\
+\sum_{i=1}^{\con n}\mathbb E_{\substack{\vct w\sim p_l(\cdot;\vct\theta_l)\\{\vct x_i}\sim p_d(\cdot;\vct\theta^{(i)}_d)}}
	\left[[1-y_i(\tilde{\vct w}\T {\vct x}_i+b)]_+ \right]\,,
	\label{eq:learner-min-2}
\end{multline}
where we are assuming the following decompositions for the mean $\vct\mu_{\vct w}=\bmat{\vct \mu_{\tilde{\vct w}}\T& \mu_{b}}\T$ and standard deviation
$\vct\sigma_{\vct w}=\bmat{\vct \sigma_{\tilde{\vct w}}\T& \sigma_{b}}\T$.
The hard part for the minimization is the term in the expectation, which can not be expressed to our knowledge in a closed-form function of the Gaussian's parameters. We thus resort 
 to a Central-Limit-Theorem-like approximation, by regarding $s_i=1-y_i(\tilde{\vct w}\T {\vct x}_i+b)$ as a Gaussian-distributed variable with mean $\mu_{s_i}$ and standard deviation $\sigma_{s_i}$, \ie{} $s_i\sim\mathcal N(\mu_{s_i},\sigma_{s_i})$. In general, $s_i$ does not follow a Gaussian distribution, since the product of two normal deviates is not normally distributed. However, if the number of features $\con k$ is large, the approximation becomes reasonable. Under this assumption, we can rewrite the expectation as follows:
\begin{align}
\nonumber
	\mathbb E_{\substack{\vct w\sim p_l(\cdot;\vct\theta_l)\\{\vct x_i}\sim p_d(\cdot;\vct\theta^{(i)}_d)}}
\left[[1-y_i(\tilde{\vct w}\T {\vct x}_i+b)]_+ \right]  \\
=\mathbb E_{s_i\sim\mathcal N(\mu_{s_i},\sigma_{s_i})}[ [s_i]_+ ]\,.
\label{eq:integral}
\end{align}
The mean and variance of the Gaussian distribution in the right-hand-side of Eq.~\eqref{eq:integral} are respectively given by 
\begin{align}
\nonumber \mu_{s_i}&=\mathbb E_{\substack{\vct w\sim p_l(\cdot;\vct\theta_l)\\{\vct x_i}\sim p_d(\cdot;\vct\theta^{(i)}_d)}}\left[1-y_i(\tilde{\vct w}\T {\vct x}_i+b) \right]  \\
&=1-y_i(\vct \mu_{\tilde{\vct w}}\T \vct\mu_{\vct x_i}+\mu_b)\,,\\
\nonumber	\sigma^{2}_{s_i}&={\mathbb V}_{\substack{\vct w\sim p_l(\cdot;\vct\theta_l)\\{\vct x_i}\sim p_d(\cdot;\vct\theta^{(i)}_d)}} 
\left[1-y_i(\tilde{\vct w}\T {\vct x}_i+b) \right] 		
\\ &={\vct\sigma_{\tilde{\vct w}}^2}\T(\vct\sigma_{{\vct x_i}}^2+\vct\mu_{{\vct x_i}}^2)+{\vct\mu_{\tilde{\vct w}}^2}\T{\vct\sigma_{{\vct x_i}}^2}+\sigma_b^2\,,
	\label{eq:sigma_si}
\end{align}
where $\mathbb V$ is the variance operator, and we assume that squaring a vector corresponds to squaring each single component.

The expectation in Eq.~\eqref{eq:integral} can be transformed after simple manipulations into the following function involving the Gauss error function (integral function of the standard normal distribution) denoted as $\erf()$:
\begin{multline}
h(\mu_{s_i},\sigma_{s_i})=\frac{\sigma_{s_i}}{\sqrt{2\pi}}\exp\left( -\frac{\mu^2_{s_i}}{2\sigma^2_{s_i}} \right)\\
	+\frac{\mu_{s_i}}{2}\left[ 1-\erf\left( -\frac{\mu_{s_i}}{\sqrt 2 \sigma_{s_i}} \right) \right]\,.
	\label{eq:h}
\end{multline}

The learner's cost in Eq.~\eqref{eq:learner-min-2} can thus be approximated as:
\begin{multline}
\label{eq:obj-linear-classifier}
\bar c_l(\vct\theta_l,\vct\theta_d)\approx L_l(\vct\mu_{{\vct w}},\vct\sigma_{{\vct w}})=
\frac{\rho_l}{2}\left( \Vert\vct \mu_{\tilde{\vct w}}\Vert^2 + \Vert\vct\sigma_{\tilde{\vct w}} \Vert^2 \right)\\
	+\sum_{i=1}^{\con n}h(\mu_{s_i}(\theta_l),\sigma_{s_i}(\theta_l))\,.
\end{multline}

We can now approximate the gradient $\nabla_{\vct\theta_l} \overline c_l$ in terms of $\nabla_{\vct\theta_l}L_l$. In the following, we denote the Hadamard (\aka{} entry-wise) product between any two vectors $\vct a$ and $\vct b$ as $\vct a \circ \vct b$, and we assume any scalar-by-vector derivative to be a column vector.
The gradients of interest are given as:
\begin{align}
\frac{\partial L_l}{\partial \vct \mu_{\vct w}} &=  \rho_l\begin{bmatrix} \vct \mu_{\tilde{\vct w}} \\ 0 \end{bmatrix}  +   \sum_{i=1}^{\con n} 
\left ( 
\frac{\partial h}{\partial \mu_{s_{i}}}\frac{\partial \mu_{s_{i}}}{\partial \vct \mu_{\vct w}}
+
\frac{\partial h}{\partial \sigma^{2}_{s_{i}}}\frac{\partial \sigma^{2}_{s_{i}}}{\partial \vct \mu_{\vct w}}
\right ) \, , \\
\frac{\partial L_l}{\partial \vct \sigma_{\vct w}}& = \rho_l \begin{bmatrix} \vct \sigma_{\tilde{\vct w}} \\ 0 \end{bmatrix}  +   \sum_{i=1}^{\con n} 
\left ( 
\frac{\partial h}{\partial \mu_{s_{i}}}\frac{\partial \mu_{s_{i}}}{\partial \vct \sigma_{\vct w}}
+
\frac{\partial h}{\partial \sigma^{2}_{s_{i}}}\frac{\partial \sigma^{2}_{s_{i}}}{\partial \vct \sigma_{\vct w}}
\right ) \, ,
\end{align}
where it is not difficult to show that
\begin{eqnarray}
\label{eq:hpart-mu}
\frac{\partial h}{\partial \mu_{s_{i}}} & =& \frac{1}{2} \left [ 1- {\rm erf}  \left (  -\frac{1}{\sqrt{2}}\frac{\mu_{s_{i}}}{\sigma_{s_{i}}} \right) \right ] \, ,\\
\label{eq:hpart-sigma}
\frac{\partial h}{\partial \sigma_{s_{i}}^{2}} &=& \frac{1}{2} \frac{1}{\sqrt{2 \pi} \sigma_{s_{i}}} \exp{\left (-\frac{1}{2} \frac{\mu_{s_{i}}^{2}}{\sigma_{s_{i}}^{2}} \right )} \, ,
\end{eqnarray}
and that
\begin{align}
\frac{\partial \mu_{s_{i}}}{\partial \vct \mu_{\vct {w}} } &= -y_{i} \begin{bmatrix} \vct \mu_{{\vct x_{i}}}\\ 1 \end{bmatrix}  , 
&&\frac{\partial \mu_{s_{i}}}{\partial \vct \sigma_{\vct {w}} } = \vct 0 \, ,\\
\frac{\partial \sigma_{s_{i}}^{2}}{\partial \vct \mu_{\vct  w}}& =  \begin{bmatrix} 2\vct \sigma^{2}_{{\vct x_{i}}} \circ \vct \mu_{\vct{\tilde  w}} \\ 0 \end{bmatrix}  , 
&&\frac{\partial \sigma_{s_{i}}^{2}}{\partial \vct \sigma_{\vct  w}} = 2\vct \sigma_{\vct  w} \circ  \begin{bmatrix} \vct\sigma_{{\vct x_i}}^2+\vct\mu_{{\vct x_i}}^2 \\ 1 \end{bmatrix}  .
\end{align}

\vspace{-10pt}
\subsection{Gradient of the data generator's cost}
In this section we turn to the data generator and we focus on approximating $\nabla_{\vct\theta_d}\overline c_d$, where $\overline c_d$ is defined as in Eq.~\eqref{eq:cd_SVM}.
We can separate $\overline c_d$ into the sum of $\con n$ functions acting on each data sample independently, \ie{} $\overline c_d(\vct\theta_l,\vct\theta_d)=\sum_{i=1}^\con n \overline c_d^{(i)}(\vct\theta_l,\vct\theta_d^{(i)})$, where for each $i\in\{1,\dots,\con n\}$:
\begin{multline}
\overline c_d^{(i)}(\vct\theta_l,\vct\theta_d^{(i)})=\mathbb E_{\substack{\vct w\sim p_l(\cdot;\vct\theta_l)\\ {\vct x_i}\sim p_d(\cdot;\vct\theta^{(i)}_d)}}\left[
	 \frac{\rho_d}{2}\Vert {\vct x_i}-\hat{\vct x_i}\Vert^2\right.\\
	 \left.+ [1+y_i(\tilde{\vct w}\T  {\vct x}_i+b)]_+ \right]\,.
	\label{eq:subprobl}
\end{multline}
By exploiting properties of the expectation and since $p_d(\cdot;\vct\theta^{(i)}_d)$ is an axis-aligned Gaussian distribution with mean $\vct \mu_{\vct x_i}$ and standard deviation $\vct \sigma_{\vct x_i}$, we can simplify Eq.~\eqref{eq:subprobl} as:
\begin{multline}
\overline c_d^{(i)}(\vct\theta_l,\vct\theta_d^{(i)})=
	\frac{\rho_d}{2}\left( \Vert\vct \mu_{\vct x_i}-\hat{\vct x_i} \Vert^2 + \Vert\vct\sigma_{\vct x_i}\Vert^2 \right) 
	\\
	+\,\mathbb E_{\substack{\vct w\sim p_l(\cdot;\vct\theta_l)\\ {\vct x_i}\sim p_d(\cdot;\vct\theta^{(i)}_d)}}
	 \left[[1+y_i(\tilde{\vct w}\T  {\vct x}_i+b)]_+ \right]\,.
	\label{eq:subprobl2}
\end{multline}
As in the case of the learner, the expectation is a troublesome term having the same form of \eqref{eq:integral}, except for an inverted sign. We adopt the same approximation used in Sect.~\ref{ssec:learner} to obtain a closed-form function. 
Accordingly, $t_i=1+y_i(\tilde{\vct w}\T  {\vct x}_i+b)$ is assumed to be normally distributed with mean $\mu_{t_i}$ and $\sigma_{t_i}$. Then
the expectation in Eq.~\eqref{eq:subprobl2} can be approximated as $h(\mu_{t_i},\sigma_{t_i})$, where function $h$ is defined as in Eq.~\eqref{eq:h}.
The variance $\sigma^{2}_{t_i}$ is equal to $\sigma^{2}_{s_i}$ (Eq.~\ref{eq:sigma_si}), while $\mu_{t_i}$ is given by:
\[
	\begin{aligned}
	\mu_{t_i}&=\mathbb E_{\substack{\vct w\sim p_l(\cdot;\vct\theta_l)\\ {\vct x_i}\sim p_d(\cdot;\vct\theta^{(i)}_d)}} 
	\left[1+y_i(\tilde{\vct w}\T  {\vct x}_i+b) \right]\\&=1+y_i(\vct \mu_{\tilde{\vct w}}\T \vct \mu_{\vct x_i}+\mu_b)\,.
	\end{aligned}
\]

The sample-wise cost of the data generator (Eq.~\ref{eq:subprobl2}) can thus be approximated as
\begin{multline}
\label{eq:obj-linear-attacker}
\overline c_d^{(i)}(\vct\theta_l,\vct\theta_d^{(i)})\approx 
 L_d(\vct\mu_{\vct x_i},\vct\sigma_{\vct x_i})=
\frac{\rho_d}{2}\left( \Vert\vct \mu_{\vct x_i}-\hat{\vct x_i} \Vert^2 + \Vert\vct\sigma_{\vct x_i}\Vert^2 \right) \\
	+\,h(\mu_{t_i}(\theta^{(i)}_d),\sigma_{t_i}(\theta^{(i)}_d))\,.
\end{multline}

The corresponding gradient is given by
\begin{align}
\frac{\partial L_d}{\partial \vct \mu_{\vct x_{i}}} &=
( \vct \mu_{\vct x_{i}}- \hat{\vct x_{i}})
+  \rho_{d} \left (
\frac{\partial h}{\partial \mu_{t_{i}}} \frac{\partial \mu_{t_{i}}}{\partial \vct \mu_{\vct x_{i}}}
+ \frac{\partial h}{\partial \sigma^{2}_{t_{i}}} \frac{\partial \sigma^{2}_{t_{i}}}{\partial \vct \mu_{\vct x_{i}}}
\right) \, , \\
\frac{\partial L_d}{\partial \vct \sigma_{\vct x_{i}}} &=
\vct \sigma_{\vct x_{i}}
+  \rho_{d} \left (
\frac{\partial h}{\partial \mu_{t_{i}}} \frac{\partial \mu_{t_{i}}}{\partial \vct \sigma_{\vct x_{i}} }
+ \frac{\partial h}{\partial \sigma^{2}_{t_{i}}} \frac{\partial \sigma^{2}_{t_{i}}}{\partial \vct \sigma_{\vct x_{i}}}
\right) \, , 
\end{align}
where $\frac{\partial h}{\partial \mu_{t_{i}}}$ and $\frac{\partial h}{\partial \sigma_{t_{i}}^{2}}$ are given as in Eqs.~\eqref{eq:hpart-mu}-\eqref{eq:hpart-sigma}, and
\begin{align}
\frac{\partial \mu_{t_{i}}}{\partial \vct \mu_{\vct x_i }} &= y_{i} \vct \mu_{\tilde{\vct w}} \, , 
&&\frac{\partial \mu_{t_{i}}}{\partial \vct \sigma_{\vct x_{i}} } = \vct 0 \, ,\\
\frac{\partial \sigma_{t_{i}}^{2}}{\partial \vct \mu_{\vct x_{i}}} &=  2\vct \sigma^{2}_{\tilde{\vct w}} \circ \vct \mu_{\vct x_{i}}
 \, , 
&&\frac{\partial \sigma_{t_{i}}^{2}}{\partial \vct \sigma_{\vct x_{i}}} = 2\vct \sigma_{\vct  x_{i}} \circ  \left ( \vct\sigma_{\tilde{\vct w}}^2+\vct\mu_{\tilde{\vct w}}^2 \right ) \, .
\end{align}

\begin{figure*}[t]
\centering
\includegraphics[width=0.9\textwidth]{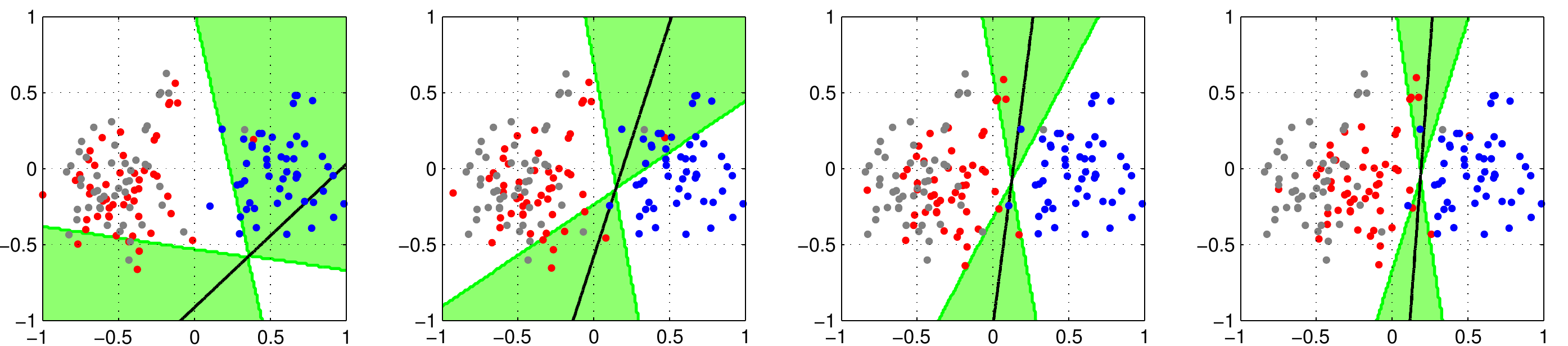}
\includegraphics[width=0.9\textwidth]{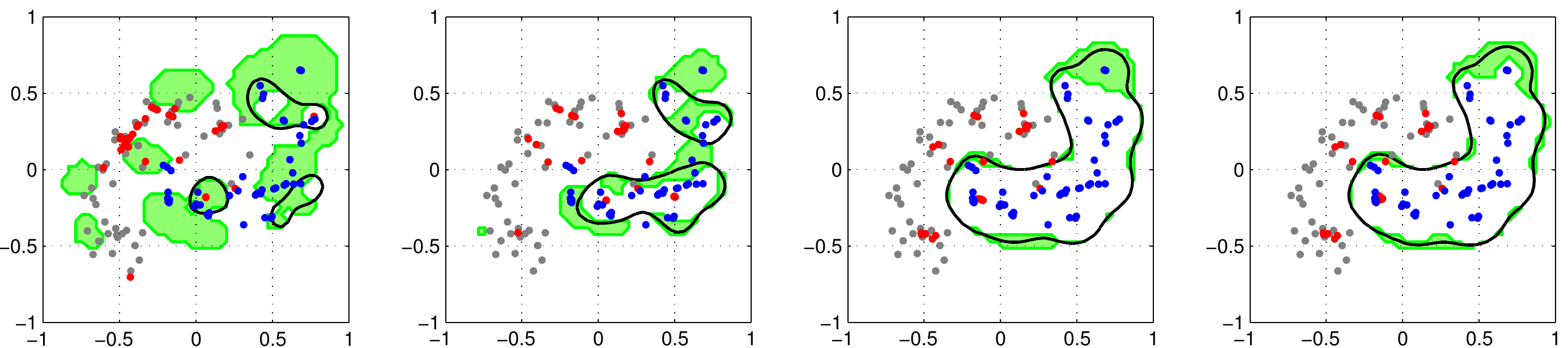}
\caption{Two-dimensional examples of randomized prediction games, for SVMs with linear (top) and RBF kernels (bottom). Each row shows how the algorithm gradually converges to a Nash equilibrium. Blue (gray) points represent the legitimate (malicious) class. The mean of each manipulated attack sample is shown as a red point (for clarity, its variance is not shown). The black solid line represents the expected decision boundary, and the green shaded area highlights its variability within one standard deviation. Note how the linear SVM's decision boundary tends to shift towards the legitimate class, while the nonlinear boundary provides a better enclosing of the same class. This intuitively allows for a higher robustness to different kinds of attack, as it requires the adversary to make a higher number of modifications to the attack samples to evade detection, at the expense of a higher number of misclassified legitimate samples.}
\label{fig:example}
\vspace{-15pt}
\end{figure*}

\section{Kernelization} \label{sec:kernelization}

Our game, as in Bruckner \etal~\cite{bruckner12}, 
assumes explicit knowledge of the feature space $\phi$, where the data generator is assumed to randomize the samples $\dot{\vct x} = \phi(\vct x)$.  However, in many applications, the feature mapping is only implicitly given in terms of a positive semidefinite \emph{kernel}  function $k:\set X\times\set X\to\mathbb R$ that measures the similarity between samples as a scalar product in the corresponding kernel Hilbert space, \ie{}, there exists $\phi:\set X\to\mathbb R$ such that $k(\vct x,\vct x')=\phi(\vct x)\T\phi(\vct x')$. Note that in this setting the input space $\set X$ is not restricted to a vector space like in the previous section (\eg{} it might contain graphs or other structured entities).

For the representer theorem to hold~\cite{scholkopf99}, 
we assume that the randomized weight vectors of the learner live in the same subspace of the reproducing kernel Hilbert space, \ie{}, $\vct w=\sum_j\alpha_j\phi(\hat{\vct x}_j)$, where $\vct\alpha\in\mathbb R^{\con n}$. Analogously, we restrict the randomized samples obtained by the data generator to live in the span of the mapped training instances, \ie{}, $\dot{\vct x}_i=\sum_{j=1}^\con n\xi_{ij}\phi(\hat{\vct x}_j)$, where $\vct \xi_i=(\xi_{i1},\dots,\xi_{i\con n})\T\in\mathbb R^{\con n}$. 

Now, instead of randomizing $\vct w$ and $\dot{\mat X}$, we let the data generator and the learner randomize $\mat\Xi=(\vct \xi_1,\dots,\vct \xi_{\con n})$ and $\vct \alpha$, respectively. Moreover, we assume 
that the expected costs $\bar c_{l/d}$ can be rewritten in terms of $\vct\alpha$ and $\mat\Xi$ in a way that involves only inner products of $\phi(\vct x)$, to take advantage of the kernel trick. This is clearly possible for the term $\vct w\T\dot{\vct x}_i=\vct \alpha\T\mat K\vct\xi_i$ in \eqref{eq:costs_l} and \eqref{eq:costs_d}, where $\mat K$ is the kernel matrix. Hence, the applicability of the kernel trick only depends on the choice of the regularizers. 
It is easy to see that due to the linearity of the variable shift, existence and uniqueness of a Nash equilibrium in our kernelized game hold under the same conditions given for the linear case.\footnote{Note that, on the contrary, manipulating samples directly in the input space would not even guarantee the existence of a Nash equilibrium, as the data generator's expected cost becomes non-quasi-convex in $\vct x$ for many (nonlinear) kernels, invalidating Theorem~\ref{prop:cond}.}

Although the data generator is virtually randomizing strategies in some subspace of the reproducing kernel Hilbert space, in reality manipulations should occur in the original input space.
Hence, to construct the real attack samples $\{\vct x_i\}_{i=1}^{n}$ corresponding to the data generator's strategy at the Nash equilibrium, 
one should solve the so-called pre-image problem, inverting the implicit feature mapping $\phi^{-1}(\mat K\vct\xi_i)$ for each sample.
This problem is in general neither convex, nor it admits a unique solution. However, reliable solutions can be easily found using well-principled approximations~\cite{scholkopf99,bruckner12}. It is finally worth remarking that solving the pre-image problem is not even required from the learner's perspective, \ie, to train the corresponding, secure classification function.

\section{Discussion} \label{sec:discussion}

In this section, we report a simple case study on a two-dimensional dataset to visually demonstrate the effect of randomized prediction games on SVM-based learners. From a pragmatic perspective, this example suggests also that uniqueness of the Nash Equilibrium should not be taken as a strict requirement in our game.

An instance of the proposed randomized prediction game for a linear SVM and for a non-linear SVM with the RBF kernel is reported in Fig.~\ref{fig:example}.
As one may note from the plots, the main effect of simulating the presence of an attacker that manipulates malicious data to evade detection is to cause the linear decision boundary to gradually shift towards the legitimate class, and the nonlinear boundary to find a better enclosing of the legitimate samples.
This should generally improve the learner's robustness to any kind of evasion attempt, as it requires the attacker to mimic more carefully the feature values of legitimate samples -- a task typically harder in several adversarial settings than just obfuscating the content of a malicious sample to make it sufficiently different from the known malicious ones~\cite{biggio13-ecml,biggio14-tkde}.

Based on this observation, any attempt aiming to satisfy the sufficient conditions for uniqueness of the Nash Equilibrium will result in an increase of the regularization strength in either the learner's or the attacker's cost function. Indeed, to satisfy the condition in Lem.~\ref{thm:new_cond}, one could sufficiently increase $\rho_l$, $\rho_d$, or both. This amounts to increasing the regularization strength of either players, which in turn reduces in some sense their power.
Hence, it should be clear that enforcing satisfaction of the sufficient conditions that guarantee the uniqueness of the Nash equilibrium might be counterproductive, by inducing the learner to weakly enclose the legitimate class, either due to a too strong regularization of the learners' parameters, 
or by limiting the ability of the attacker to manipulate the malicious samples, thus allowing the leaner to keep a loose boundary. 
This will in general compromise the quality of the adversarial learning procedure.
This argument shares similarities with the idea of addressing non-convex machine learning problems directly, without resorting to convex approximations~\cite{collobert06,bengio07}.

Besides improving classifier robustness, finding a better enclosure of the legitimate class may however cause a higher number of legitimate samples to be misclassified as malicious. There is indeed a trade-off between the desired level of robustness and the fraction of misclassified legitimate samples.
The benefit of using randomization here is to make the attacker's strategy less pessimistic than in the case of static prediction games~\cite{bruckner09,bruckner12}, which should allow us to eventually find a better trade-off between robustness and legitimate misclassifications. 
This aspect is investigated more systematically in the experiments reported in the next section.

\section{Experiments} \label{sec:exp}

\begin{figure*}[tbp]
\centering
\includegraphics[width=0.24\textwidth]{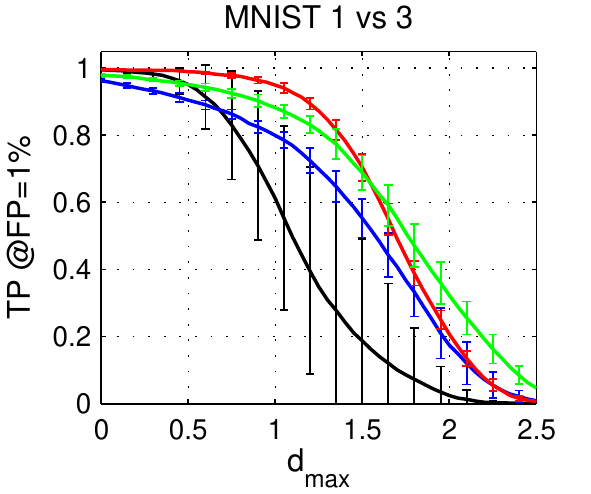}
\includegraphics[width=0.24\textwidth]{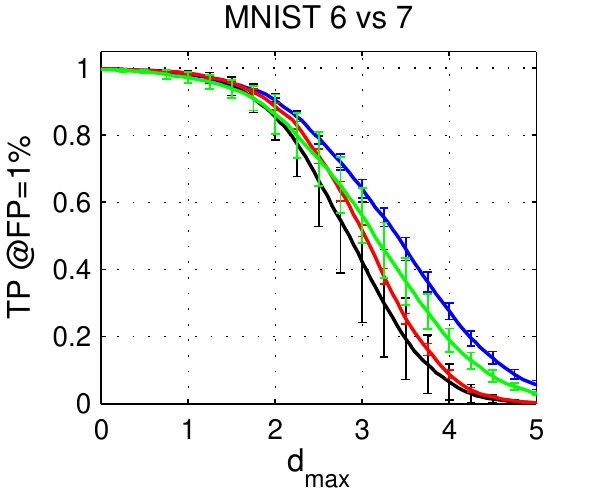}
\includegraphics[width=0.24\textwidth]{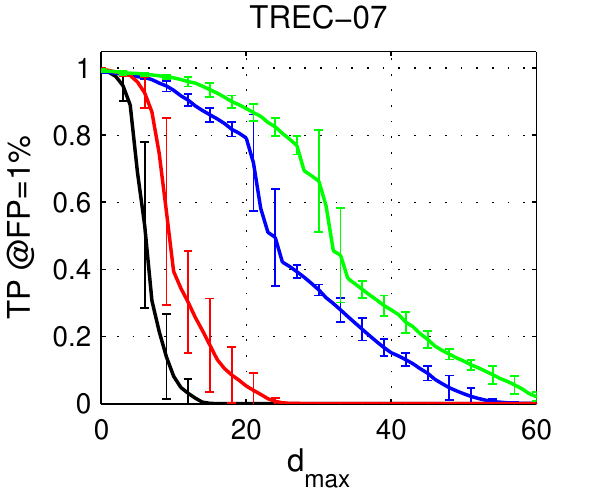}
\includegraphics[width=0.24\textwidth]{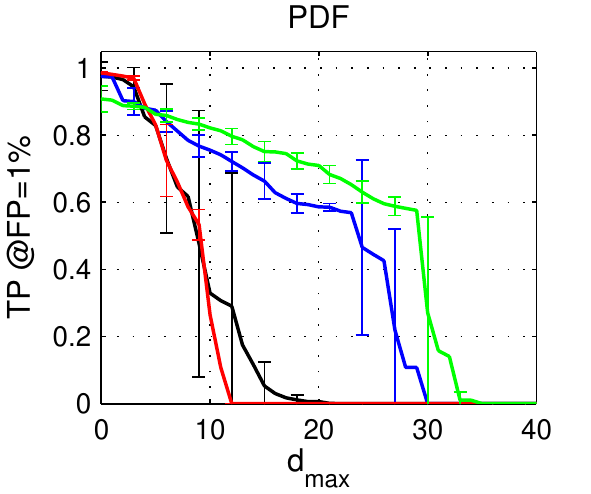}\\
\includegraphics[width=0.5\textwidth]{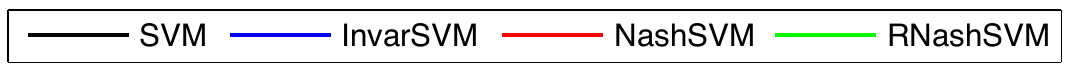} \vspace{-5pt}
\caption{Security evaluation curves, reporting the average TP at FP=1\% (along with its standard deviation, shown with error bars) against an increasing amount of manipulations to the attack samples (measured by $d_{\rm max}$), for handwritten digit (first and second plot), spam (third plot), and PDF (fourth plot) data.}
\label{fig:exp-results}
\end{figure*}

In this section we present a set of experiments on handwritten digit recognition, spam filtering, and PDF malware detection.
Despite handwritten digit recognition is not a proper adversarial learning task as spam and malware detection, we consider it in our experiments to provide a \emph{visual} interpretation of how secure learning algorithms are capable of improving robustness to evasion attacks.

We consider only linear classifiers, as they are a typical choice in these settings, and especially in spam filtering~\cite{lowd05,biggio10-ijmlc,biggio14-tkde}. This also allows us to carry out a fair comparison with state-of-the-art secure learning algorithms, as they yield linear classification functions.
We compare our secure linear SVM learner (Sect.~\ref{sec:case_study}) with the standard linear SVM implementation~\cite{libsvm}, and with the state-of-the-art robust classifiers InvarSVM~\cite{globerson06-icml,teo08}, and NashSVM~\cite{bruckner12} (see Sect.~\ref{sec:related-work}). 

The goal of these experiments is to test whether these secure  algorithms work well also under attack scenarios that differ from those hypothesized during design -- a typical setting in security-related tasks; \eg, what happens if game-based classification techniques like that proposed in this paper and NashSVM are used against attackers that exploit a different attack strategy, \ie{}, attackers that may not act rationally according to the hypothesized objective function? What happens when the attacker does not play at the expected Nash equilibrium? These are rather important questions to address, as we do not have any guarantee that real-world attackers will play according to the hypothesized objective function.

\textbf{Security evaluation.} To address the above issues, we consider the security evaluation procedure proposed in~\cite{biggio14-tkde}. It evaluates the performance of the considered classifiers under attack scenarios of increasing \emph{strength}.
We consider the True Positive (TP) rate (\ie, the fraction of detected attacks) evaluated at $1\%$ False Positive (FP) rate (\ie, the fraction of misclassified legitimate samples) as performance measure.
We evaluate the performance of each classifier in the absence of attack, as in standard performance evaluation techniques, and then start manipulating the malicious test samples to simulate attacks of different strength.
We assume a worst-case adversary, \ie{}, an adversary that has perfect knowledge of the attacked classifier, since we are interested in understanding the worst-case performance degradation. Note however that other choices are possible, depending on specific assumptions on the adversary's knowledge and capability~\cite{biggio10-ijmlc,biggio14-tkde,kolcz09}.
In this setting, we assume that the optimal (worst-case) sample manipulation $\mathbf x^{*}$ operated by the attacker is obtained by solving the following optimization problem:
\begin{eqnarray}
\label{eq:sec-eval}
\begin{array}{rl}
\vct x^{*} \in \displaystyle\argmin_{\vct x}& y  f(\vct x;\vct w),\\
{\rm s.t.}&d(\vct x, \hat{\vct x_{i}}) \leq d_{\rm max},
\end{array}
\end{eqnarray}
where $y$ is the malicious class label, $d(\vct x, \vct x_{i})$ measures 
the distance between the perturbed sample $\vct x$ and the $i^{\rm th}$ malicious data sample $\hat{\vct x_i}$ (in this case, we use the $\ell_{2}$ norm, as done by the considered classifiers). 
The maximum amount of modifications is bounded by $d_{\rm max}$, which is a parameter representing the attack strength. It is obvious that the more modifications the adversary is allowed to make on the attack samples, the higher the performance degradation incurred by the classifier is expected to be. Accordingly, the performance of more secure classifiers is expected to degrade more gracefully as the attack strength increases~\cite{biggio10-ijmlc,biggio14-tkde}.

The solution of the above problem is trivial when we consider linear classifiers, the Euclidean distance, and $\mathbf x$ is unconstrained: it amounts to setting $\vct x^{*} = \hat{\vct x_{i}} - y d_{\rm max} \frac{\mathbf w}{||\mathbf w||}$. If $\mathbf x$ lies within some constrained domain, \eg{} $[0,1]$, then one may consider a simple gradient descent with box constraints on $\mathbf x$ (see, \eg,~\cite{biggio13-ecml}).
If $\mathbf x$ takes on binary values, \eg{}, $\{0,1\}$, then the attack amounts to switching from $0$ to $1$ or vice-versa the value of a maximum of $d_{\max}$ features which have been assigned the highest absolute weight values by the classifier. In particular, if $y\,w_{k} > 0$ ($y\,w_{k} < 0$) and the $k$-th feature satisfies $\hat{x}_{ik}=1$ ($\hat{x}_{ik}=0$), then $x^{*}_{k}=1$ ($x^{*}_{k}=0$)~\cite{biggio10-ijmlc,biggio14-tkde}. 

\textbf{Parameter selection.} The considered methods require setting different parameters. From the learners' perspective, we have to tune the regularization parameter $C$ for the standard linear SVM and InvarSVM, while we respectively have $\rho_{-1}$ and $\rho_{l}$ for NashSVM and for our method. 
In addition, the robust classifiers require setting the parameters of their attacker's objective.
For InvarSVM, we have to set $K$, \ie{}, the number of modifiable features, while for NashSVM and for our method, we have to set the value of the regularization parameter $\rho_{+1}$ and $\rho_{d}$, respectively.
Further, to guarantee existence of a Nash Equilibrium point, we have to enforce some box constraints on the distribution's parameters. For the attacker, we restrict the mean of the attack points to lie in $[0,1]$ (as the considered datasets are normalized in that interval), and their variance in $[10^{-3},0.5]$. For the learner, the variance of $\vct w$ is allowed to vary in  $[10^{-6},10^{-3}]$, while its mean takes values on $[-W,W]$, where $W$ is optimized together with the other parameters.
All the above mentioned parameters are set by performing a grid-search on the parameter space
$( C$, $\rho_{-1}$, $\rho_{d} \in \{0.01, 0.1, 1, 10, 100\}$;
$K \in \{ 8, 13, 25, 30, 47 , 52,  63\}$;
$\rho_{+1}, \rho_{d} \in \{0.01, 0.05, 0.1, 1, 10\}$;
$W \in \{0.01, 0.05, 0.1, 1\} )$, and retaining the parameter values that maximize the area under the security evaluation curve on a validation set.
The reason is to find a parameter configuration for each method that attains the best average robustness over all attack intensities (values of $d_{\rm max}$), \ie{} the best average TP rate at FP=$1\%$.

\vspace{-10pt}
\subsection{Handwritten Digit Recognition}
Similarly to \cite{globerson06-icml}, we focus on two two-class sub-problems of discriminating between two
distinct digits from the MNIST dataset \cite{LeCun95}, \ie{} 1 vs 3, and 6 vs 7,
where the second digit in each pair represents the attacking class ($y=+1$).
The digits are originally represented as gray-scale images of $28 \times 28$ pixels.
They are simply mapped to feature vectors by ordering the pixels in raster scan order.
The overall number of features is thus $d = 784$.  We normalize each feature (pixel value) in $[0,1]$,
by dividing its value by $255$.
We build a training and a validation set of 1,000 samples each by randomly sampling the original training data available for MNIST. As for the test set, we use the default set provided with this data, which consists of approximately 1,000 samples for each digit (class).
The results averaged on $5$ repetitions are shown in the first and second plot of Fig.~\ref{fig:exp-results} respectively for 1 vs 3, and 6 vs 7. 
As one may notice, in the absence of attack (\ie{} when $d_{\rm max}=0$), all classifiers achieved comparable performance (the TP rate is almost 100\% for all of them), due to a conservative choice of the operating point (FP=$1\%$), that should also guarantee a higher robustness against the attack.
In the presence of attack, our approach (\emph{RNashSVM}) exhibits comparable performance to NashSVM on the problem of discriminating 1 vs 3, and to InvarSVM on 6 vs 7. NashSVM outperforms the standard SVM implementation in both cases, but exhibits lower security (robustness) than InvarSVM on 6 vs 7, despite the attacker's regularizer in InvarSVM is not even based on the $\ell_{2}$ norm.

\begin{figure}[t]
\centering
\includegraphics[width=0.968\columnwidth]{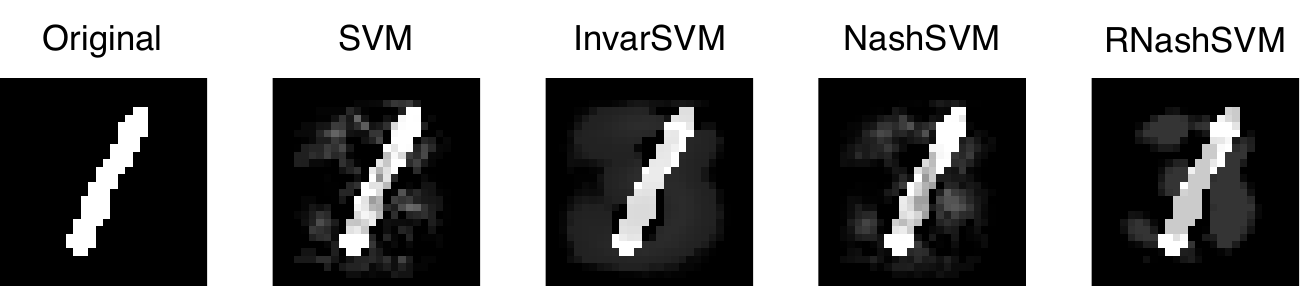}\\
\includegraphics[width=0.972\columnwidth]{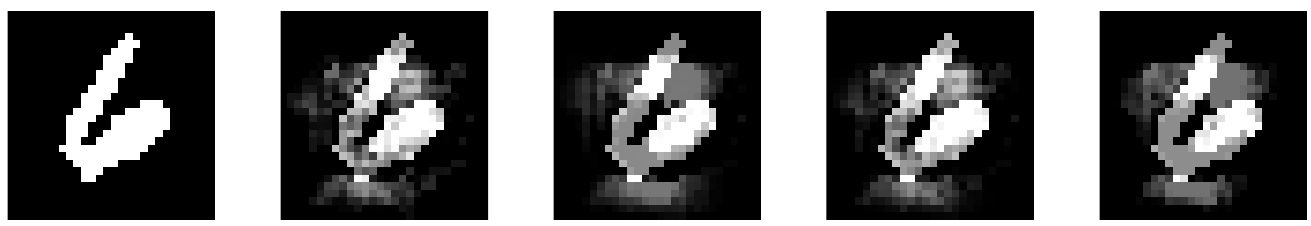} \vspace{-5pt}
\caption{Examples of obfuscated digits against each classifier when $d_{\rm max}=2.5$ for 1 vs 3 (top row), and when $d_{\rm max}=5$ for 6 vs 7 (bottom row).}
\label{fig:digits}
\end{figure}

Finally, in Fig.~\ref{fig:digits} we report two attack samples (a digit from class 1 and one from class 6) and show how they are obfuscated by the attack strategy of Eq.~\eqref{eq:sec-eval} to evade detection against each classifier. Notice how the original attacking samples (1 and 6) tend to resemble more the corresponding attacked classes (3 and 7) when natively robust classifiers are used. This visual example confirms the analysis of Sect.~\ref{sec:discussion}, \ie, that higher robustness is achieved when the adversary is required to mimic the feature values of samples of the legitimate class, 
instead of slightly modifying the attack samples to differentiate them from the rest of the malicious data.

\vspace{-10pt}
\subsection{Spam Filtering}

In these experiments we use the benchmark, publicly available, TREC 2007 email corpus \cite{trec07}, which consists of 75,419 real emails (25,220 legitimate and 50,199 spam messages) collected between April and July 2007.
We exploit the bag-of-words feature model, in which each binary feature denotes the absence ($0$) or presence ($1$) of the corresponding word in a given email \cite{kolcz09,biggio10-ijmlc,biggio14-tkde,bruckner12}.
Features (words) are extracted from training emails using the tokenization method of the widely-known anti-spam filter SpamAssassin,\footnote{\url{http://spamassassin.apache.org}}
 and then $n = 1,000$ distinct features are selected using a supervised feature selection approach based on the information gain criterion~\cite{brown12}.
We build a training and a validation set of 1,000 samples each by randomly sampling the first 5,000 emails in chronological order of the original dataset, while a test set of about 2,000 samples is randomly sampled from the subsequent set of 5,000 emails. The results averaged on $5$ repetitions are shown in the third plot of Fig.~\ref{fig:exp-results}. 
As in the previous case, in the absence of attack ($d_{\rm max}=0$) all the classifiers exhibit a very high (and similar) performance. However, as the attack intensity ($d_{\rm max}$) increases, their performance degrades more or less gracefully, \ie{} their robustness to the attack is different. Surprisingly, one may notice that only InvarSVM and RNashSVM exhibited an improved level of security.
The reason is that these two classifiers are able to find a more uniform set of weights than SVM and NashSVM, and, in this case, this essentially requires the adversary to manipulate a higher number of features to significantly decrease the value of the classifier's discriminant function. Note that a similar result has been heuristically found also in~\cite{kolcz09,biggio10-ijmlc}.

\vspace{-10pt}
\subsection{PDF Malware Detection}
We consider here another relevant application example in computer security, \ie, the detection of malware in PDF files. 
The main reason behind the diffusion of malware in PDF files is that they exhibit a very flexible structure that allows embedding several kinds of resources, including \texttt{Flash}, \texttt{JavaScript} and even executable code.  
Resources simply consists of \emph{keywords} that denote their type, and of \emph{data streams} that contain the actual object; \eg, an embedded resource in a PDF file may be encoded as follows:%
\begin{alltt}
13 0 obj << \textbf{/Kids} [ 1 0 R 11 0 R ]
\textbf{/Type /Page} ... >> end obj
\end{alltt}
where keywords are highlighted in bold face.
Recent work has exploited machine learning techniques to discriminate between malicious and legitimate PDF files, based on the analysis of their structure and, in particular, of the embedded keywords~\cite{smutz12,maiorca12-mldm,maiorca13-asiaccs,srndic13-ndss,srndic14}. 
We exploit here a similar feature representation to that proposed in \cite{maiorca12-mldm}, where each feature denotes the presence of a given keyword in the PDF file.
We collected 5993 recent malware samples from the \emph{Contagio} dataset,\footnote{\url{http://contagiodump.blogspot.it}} and 5951 benign samples from the web.
Following the procedure described in \cite{maiorca12-mldm}, we extracted 114 keywords from the first 1,000 samples (in chronological order) to build our feature set.
Then, we build training, validation and test sets as in the spam filtering case, and average results over $5$ repetitions. Attacks in this case are simulated by allowing the attacker only to increase the feature values of malicious samples, which corresponds to adding the constraint $\vct x \geq \hat{\vct x}_{i}$ (where the inequality holds for all features) to Problem~\ref{eq:sec-eval}. The reason is that removing objects (and keywords) from malicious PDFs may compromise the intrusive nature of the embedded exploitation code, whereas adding objects can be easily done through the PDF versioning mechanism~\cite{maiorca13-asiaccs,biggio13-ecml,srndic14}.

Results are shown in the $4$th plot of Fig.~\ref{fig:exp-results}. The considered methods mostly exhibit the same behavior shown in the spam filtering case, besides the fact that, here, there is a clearer trade-off between the performance in the absence of attack, and robustness under attack. In particular, InvarSVM and RNashSVM are significantly more robust under attack (\ie, when $d_{\rm max} > 0$) than SVM and NashSVM, at the expense of a slightly worsened detection rate in the absence of attack (\ie, when $d_{\rm max} = 0$).

To summarize, the reported experiments show that, even if the attacker does not play the expected attack strategy at the Nash equilibrium, most of the proposed state-of-the-art secure classifiers are still able to outperform classical techniques, and, in particular, that the proposed RNashSVM classifier may guarantee an even higher level of robustness. Understanding how this property relates to the use of probability distributions over the set of the classifier's and of the attacker's strategies remains an interesting future question. 

\section{Related Work} \label{sec:related-work}

The problem of devising secure classifiers against different kinds of manipulation of samples at test time has been widely investigated in previous work \cite{dalvi04,globerson06-icml,teo08,dekel10,huang11,bruckner09,bruckner11,bruckner12,grosshans13,vamvoudakis14,zhang15}.
Inspired by the seminal work by Dalvi \etal~\cite{dalvi04}, several authors have proposed a variety of modifications to existing learning algorithms to improve their security against different kinds of attack. 
Globerson~\etal~\cite{globerson06-icml,teo08} have formulated the so-called \emph{Invariant} SVM (InvarSVM) in terms of a \emph{minimax} approach (\ie, a zero-sum game) to deal with worst-case feature manipulations at test time, including feature addition, deletion, and rescaling. This work has been further extended in \cite{dekel10} to allow features to have different a-priori importance levels, instead of being manipulated equally likely.
Notably, more recent research has also considered the development of secure learning algorithms based on zero-sum games for sensor networks, including distributed secure algorithms~\cite{zhang15} and algorithms for detecting adversarially-corrupted sensors~\cite{vamvoudakis14}.

The rationale behind shifting from zero-sum to non-zero-sum games for adversarial learning is that the classifier and the attacker may not necessarily aim at maximizing antagonistic objective functions. This in turn implies that modeling the problem as a zero-sum game may lead one to design overly-pessimistic classifiers, as pointed out in~\cite{bruckner12}. 
Even considering a non-zero-sum Stackelberg game may be too pessimistic, since the attacker (follower) is supposed to move after the classifier (leader), while having full knowledge of the chosen classification function (which again is not realistic in practical settings)~\cite{bruckner11,bruckner12}.
For these reasons, Br\"uckner et al.~\cite{bruckner09,bruckner12} have formalized adversarial learning as a non-zero-sum game, referred to as \emph{static prediction game}. 
Assuming that the players act \emph{simultaneously} (conversely to Stackelberg games~\cite{bruckner11}), they devised conditions under which a unique \emph{Nash equilibrium} for this game exists, and developed algorithms for learning the corresponding robust classifiers, including the so-called NashSVM. Our work essentially extends this approach by introducing randomization over the players' strategies. 

For completeness, we also mention here that in \cite{grosshans13} Bayesian games for adversarial regression tasks have been recently proposed. In such games, uncertainty on the objective function's parameters of either player is modeled by considering a probability distribution over their possible values. To the best of our knowledge, this is the first attempt towards modeling the uncertainty of the attacker and the classifier on the opponent's objective function.

\section{Conclusions and Future Work} \label{sec:conclusions}
In this paper, we have extended the work in \cite{bruckner12} by introducing randomized prediction games.
To operate this shift, we have considered parametrized, bounded families of probability distributions defined over the set of pure strategies of either players.
The underlying idea, borrowed from \cite{biggio08-spr,barreno06-asiaccs,huang11}, consists of randomizing the classification function to make the attacker select a less effective attack strategy.
Our experiments, conducted on an handwritten digit recognition task and on realistic application examples involving spam and malware detection, show that competitive, secure SVM classifiers can be learnt using our approach, even when the conditions behind uniqueness of the Nash equilibrium may not hold, \ie{}, when the attacker may not play according to the objective function hypothesized for her by the classifier.
This mainly depends on the particular kind of decision function learnt by the learning algorithm under our game setting, which tends to find a better `enclosing' of the legitimate class. This generally requires the attacker to make more modifications to the malicious samples to evade detection, regardless of the attack strategy chosen. 
We can thus argue that the proposed methods exhibit robustness properties particularly suited to adversarial learning tasks.
Moreover, the fact that the proposed methods may perform well also when the Nash equilibrium is not guaranteed to be unique suggests us that the conditions behind its uniqueness may hold under less restrictive assumptions (\eg, when the SVM admits a unique solution~\cite{abe02,burges99}). We thus leave a deeper investigation of this aspect to future work. 

Another interesting extension of this work may be to apply randomized prediction games in the context of unsupervised learning, and, in particular, clustering algorithms.
It has been recently shown that injecting a small percentage of well-crafted \emph{poisoning} attack samples into the input data may significantly subvert the clustering process, compromising the subsequent data analysis~\cite{biggio13-aisec,biggio14-spr}. In this respect, we believe that randomized prediction games may help devising secure countermeasures to mitigate the impact of such attacks; \eg, by explicitly modeling the presence of poisoning samples (generated according to a probability distribution chosen by the attacker) during the clustering process.

It is worth finally mentioning that our work is also slightly related to previous work on \emph{security games}, in which the goal of the defender is to adopt randomized strategies to protect his or her assets from the attacker, by allocating a limited number of defensive resources; \eg, police officers for airport security, protection mechanisms for network security~\cite{korzhyk11,alpcan10,tambe11}.
Although our game is not directly concerned to the protection of a given set of assets, we believe that investigating how to bridge the proposed approach within this well-grounded field of study may provide promising research directions for future work, \eg, in the context of network security~\cite{alpcan10,tambe11}, or for suggesting better user attitudes towards security issues~\cite{grossklags08}. 
This may also suggest interesting theoretical advancements; \eg, to establish conditions for the equivalence of Nash and  Stackelberg games~\cite{korzhyk11}, and to address issues related to the uncertainty on the players' strategies, or on their (sometimes bounded) rationality, \eg, through the use of Bayesian games~\cite{grosshans13}, security strategies and robust optimization~\cite{alpcan10,tambe11}. 
Another suggestion to overcome the aforementioned issues is to exploit higher-level models of the interactions between attackers and defenders in complex, real-world problems; \eg, through the use of replicator equations to model adversarial dynamics in security-related tasks~\cite{cybenko12}.
Exploiting conformal prediction may be also an interesting research direction towards improving current adversarial learning systems \cite{wechsler15}.
To conclude, we believe these are all relevant research directions for future work.



\begin{biography}[{\includegraphics[width=1in,height=1.25in,clip,keepaspectratio]{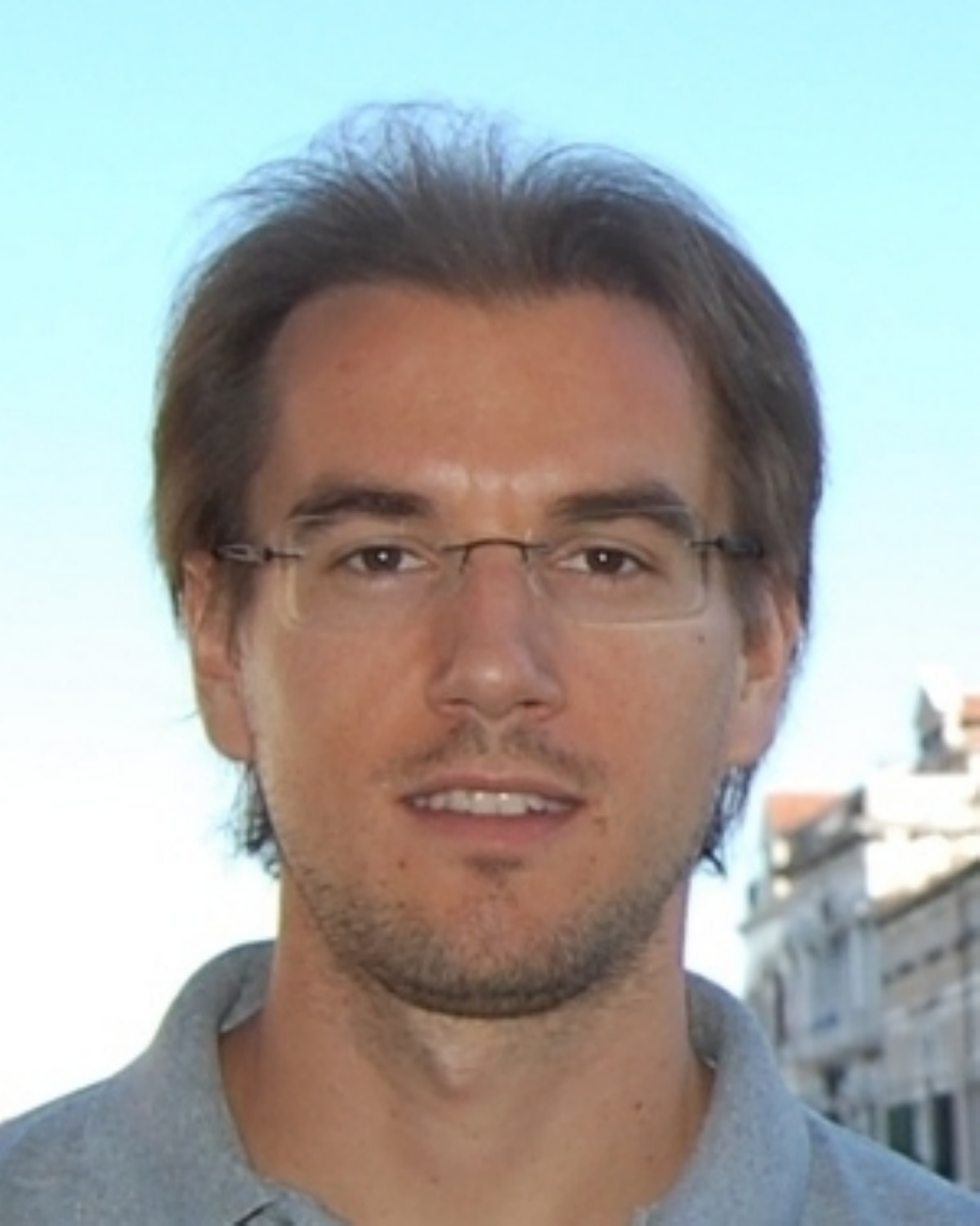}}]{Samuel Rota Bul\`o} 
received his PhD in computer science at the University of Venice, Italy, in 2009 
and he worked as a postdoctoral researcher at the same institution until 2013. He is currently a researcher of the ``Technologies of Vision'' laboratory
at Fondazione Bruno Kessler in Trento, Italy.
His main research 
interests are in the areas of computer vision and pattern recognition with particular emphasis on 
discrete and continuous optimisation methods, graph theory and game theory. Additional research 
interests are in the field of stochastic modelling.
He regularly publishes his research in well-recognized 
conferences and top-level journals mainly in the areas of computer vision and pattern recognition.
He held research visiting positions at the following institutions: IST - Technical University of Lisbon, University of Vienna, 
Graz University of Technology, University of York (UK), Microsoft Research Cambridge (UK) and University of Florence.
\end{biography}

\vspace{-0.5cm}
\begin{IEEEbiography}
[{\includegraphics[width=1in,height =1.25in,clip,keepaspectratio]{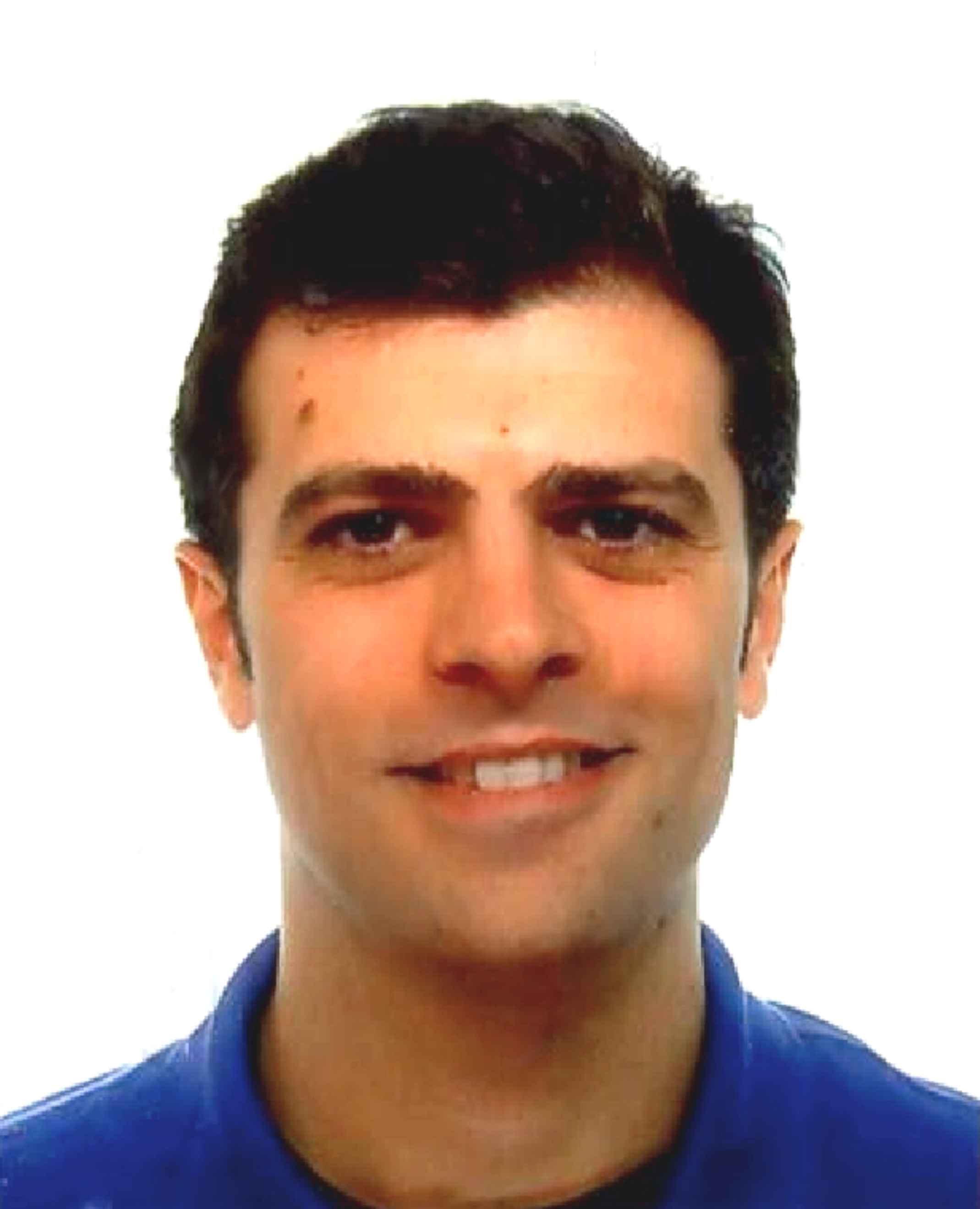}}]{Battista Biggio (M'07)} received the M.Sc. degree (Hons.) in Electronic Engineering and the Ph.D. degree in Electronic Engineering and Computer Science from the University of Cagliari, Italy, in 2006 and 2010. Since 2007, he has been with the Department of Electrical and Electronic Engineering, University of Cagliari, where he is currently a post-doctoral researcher. In 2011, he visited the University of T\"ubingen, Germany, and worked on the security of machine learning to training data poisoning. His research interests include secure machine learning, multiple classifier systems, kernel methods, biometrics and computer security. Dr. Biggio serves as a reviewer for several international conferences and journals. He is a member of the IEEE and of the IAPR.
\end{IEEEbiography}

\vspace{-0.5cm}
\begin{biography}
[{\includegraphics[width=1in,height =1.25in,clip,keepaspectratio]{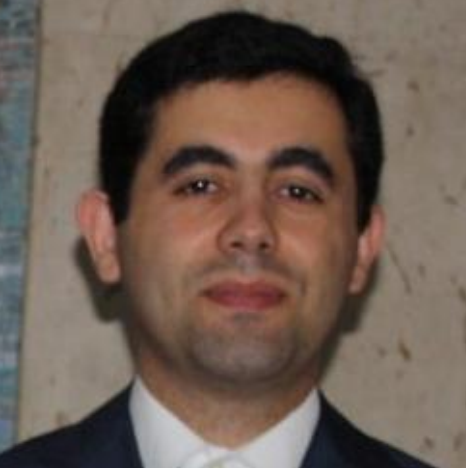}}]{Ignazio Pillai} received the M.Sc. degree in Electronic Engineering, with honors, and the Ph.D. degree in Electronic Engineering and Computer Science from the University of Cagliari, Italy, in 2002 and 2007, respectively.
Since 2003 he has been working for the Department of Electrical and Electronic Engineering at the University of Cagliari, Italy, where he is a post doc in the research laboratory on pattern recognition and applications.
His main research topics are related to multi-label classification, multimedia document categorization and classification with a reject option. He has published about twenty papers in international journals and conferences, and acts as a reviewer for several international conferences and journals.
\end{biography}

\vspace{-0.5cm}
\begin{biography}
[{\includegraphics[width=1in,height=1.25in,clip,keepaspectratio]{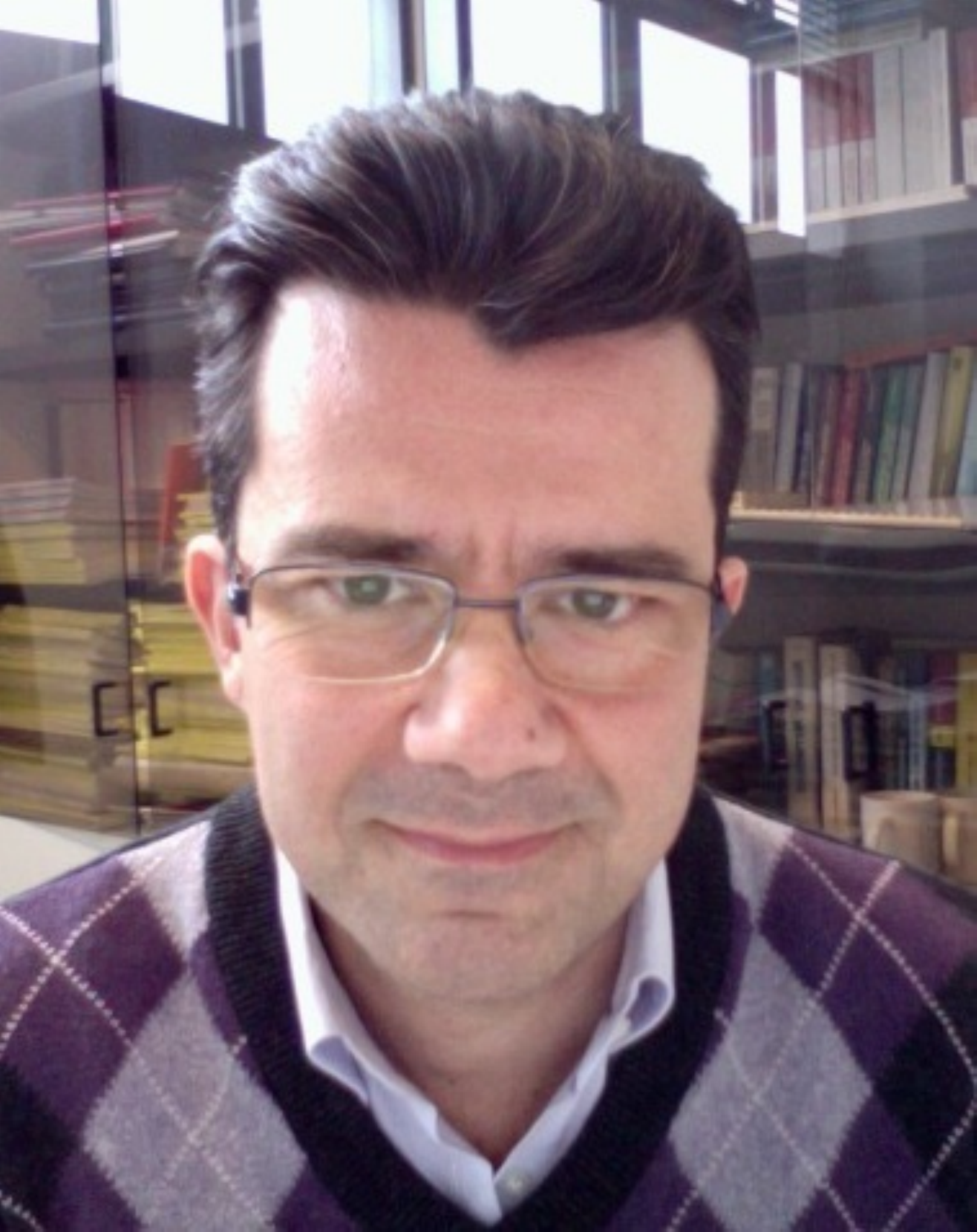}}]{Marcello Pelillo}
joined the faculty of the University of Bari, Italy, as an Assistant professor of 
computer science in 1991. Since 1995, he has been with the University of Venice, Italy, where he is 
currently a Full Professor of Computer Science. He leads the Computer Vision and Pattern Recognition 
Group and has served from 2004 to 2010 as the Chair of the board of studies of the Computer Science 
School. He held visiting research positions at Yale University, the University College London, 
McGill University, the University of Vienna, York University (UK), and the National 
ICT Australia (NICTA). He has published more than 130 technical papers in refereed journals, handbooks, 
and conference proceedings in the areas of computer vision, pattern recognition and neural computation. 
He serves (or has served) on the editorial board for the journals \emph{IEEE Transactions on Pattern 
Analysis and Machine Intelligence}, \emph{Pattern Recognition} and \emph{IET Computer Vision}, and 
is regularly on the program committees 
of the major international conferences and workshops of his fields. In 1997, he co-established a new 
series of international conferences devoted to energy minimization methods in computer vision and 
pattern recognition (EMMCVPR). He is (or has been) scientific 
coordinator of several research projects, including SIMBAD, an EU-FP7 project devoted to 
similarity-based pattern analysis and recognition. Prof. Pelillo is a Fellow of the IAPR and Fellow
of the IEEE. 
\end{biography}

\begin{IEEEbiography}[{\includegraphics[width=1in,height =1.25in,clip,keepaspectratio]{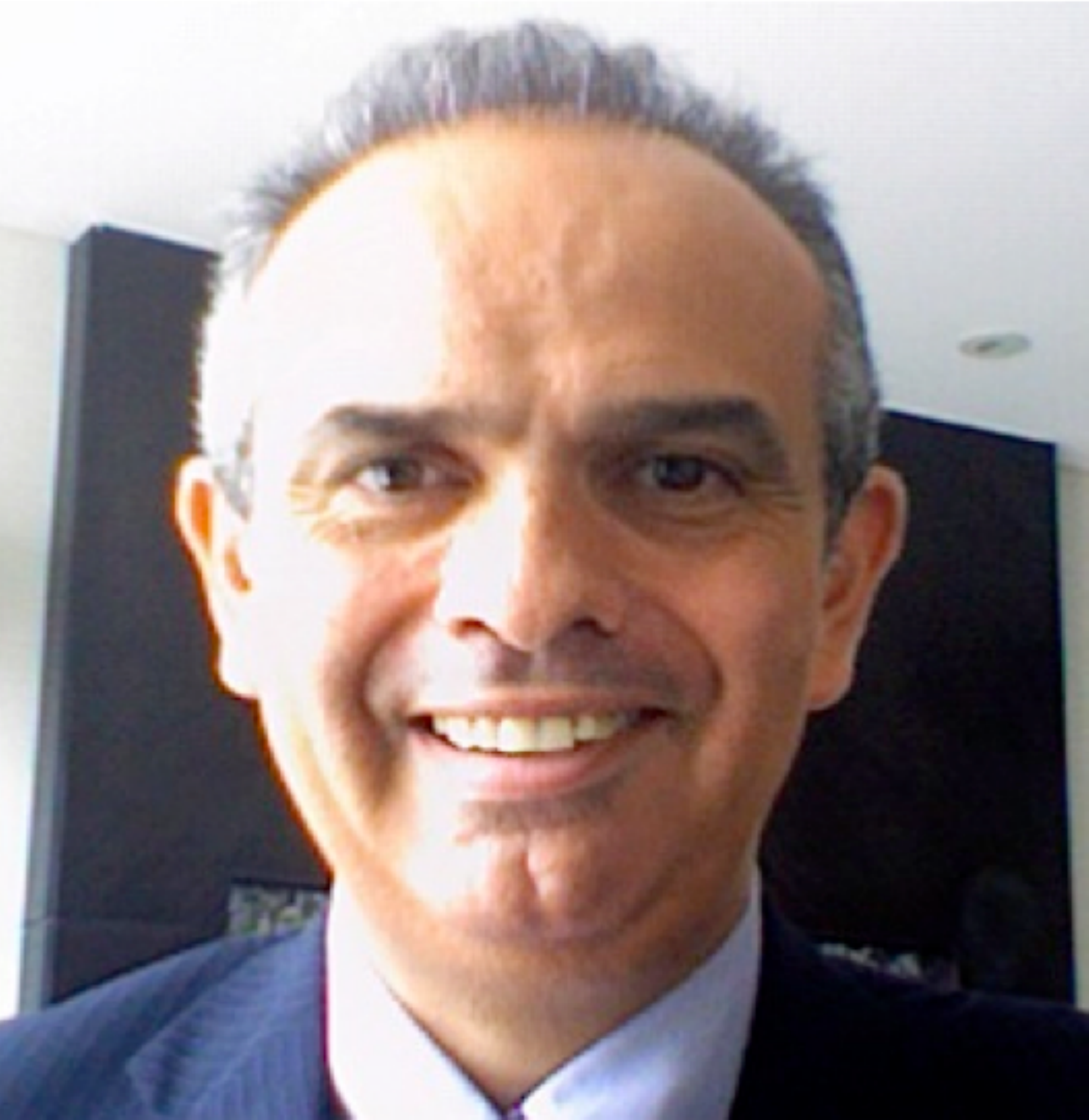}}]{Fabio Roli (F'12)}  received his Ph.D. in Electronic Engineering from the University of Genoa, Italy. He was a research group member of the University of Genoa (88-94). He was adjunct professor at the University of Trento ('93-'94). In 1995, he joined the Department of Electrical and Electronic Engineering of the University of Cagliari, where he is now professor of Computer Engineering and head of the research laboratory on pattern recognition and applications. His research activity is focused on the design of pattern recognition systems and their applications. He was a very active organizer of international conferences and workshops, and established the popular workshop series on multiple classifier systems. Dr. Roli is Fellow of the IEEE and of the IAPR.
\end{IEEEbiography}

\end{document}


\title{Supplementary Material\\\vspace{5pt}
    \Large{Randomized Prediction Games for Adversarial Machine Learning}}

\author{Samuel~Rota~Bul\`o,~\IEEEmembership{Member,~IEEE,}
Battista~Biggio,~\IEEEmembership{Member,~IEEE,}
Ignazio~Pillai,~\IEEEmembership{Member,~IEEE,}
Marcello~Pelillo,~\IEEEmembership{Fellow,~IEEE}
Fabio~Roli,~\IEEEmembership{Fellow,~IEEE}
\thanks{S. Rota Bul\`o is with ICT-Tev, Fondazione Bruno Kessler, Trento, Italy}
	\thanks{M. Pelillo is with DAIS, Universit\`a Ca' Foscari, Venezia, Italy}
	\thanks{B. Biggio, I. Pillai, F. Roli are with DIEE, University of Cagliari, Italy}}

\IEEEcompsoctitleabstractindextext{
\begin{abstract}
    This document provides sufficient conditions for the uniqueness of a Nash equilibrium in randomized prediction games. The provided conditions generalize the ones given in \cite{bruckner12}. 
\end{abstract}
}
\maketitle

\IEEEdisplaynotcompsoctitleabstractindextext

This document is devoted to provide fine-grained assumptions
that guarantee the positive definiteness of $\overline J_{\vct r}$ and, hence,
the uniqueness of the Nash equilibrium via Thm.~\ref{thm:unique_jr}.
In particular, we will provide a first set of conditions to rewrite the pseudo-Jacobian $\overline J_{\vct r}(\vct\theta_l,\vct\theta_d)$ of our game in terms of the pseudo-Jacobian $J_{\vct r}(\vct w,\dot{\mat X})$ of the underlying prediction game, \ie,
\begin{equation}
  J_{\vct r}(\vct w,\dot{\mat X})=\bmat{r_l\nabla^2_{\vct w,\vct w}c_l(\vct w,\dot{\mat X})&r_l\nabla^2_{\vct w,\dot{\mat X}}c_l(\vct w,\dot{\mat X})\\
  r_d\nabla^2_{\dot{\mat X},\vct w} c_d(\vct w,\dot{\mat X})&r_d\nabla^2_{\dot{\mat X},\dot{\mat X}}c_d(\vct w,\dot{\mat X})}\,,
  \label{eq:Jr_decomp}
\end{equation}
and then provide a further set of conditions on $J_{r}(\vct w,\dot{\mat X})$, adapted from~\cite{bruckner12}, to ensure the uniqueness of the Nash equilibrium in our game.

The first set of conditions is related to the parametrized probability distributions $p_{l/d}(\cdot;\vct\theta_{l/d})$ of the two players.
\begin{assumption}\label{ass2}
    There exist random matrices $\mat V_l\in\mathbb R^{\con m\times\con s_l}$ and $\mat V_d\in\mathbb R^{\con n\con m\times\con s_d}$ (each defined on a probability space) such that
  \begin{itemize}
    \item random variable $\vct w$ distributed as $p_l(\vct w;\vct\theta_l)$ is equivalent in distribution to $\mat V_l\vct\theta_l$ for all $\vct\theta_l\in\Theta_l$, 
    \item random variable $\dot{\mat X}$ distributed as $p_d(\dot{\mat X};\vct\theta_d)$ is equivalent in distribution to $\mat V_d\vct\theta_d$ for all $\vct\theta_d\in\Theta_d$, and
    \item random variables $\mat V_{l/d}$ do not depend on $\vct\theta_{l/d}$, respectively.
  \end{itemize}
\end{assumption}

Intuitively, random variables $\vct w$ and $\dot{\mat X}$ depend on the parameters $\vct\theta_l$ and $\vct\theta_d$ in a non-linear way via their probability distributions $p_l(\vct w;\vct\theta_l)$ and $p_d(\dot{\mat X};\vct\theta_d)$. Assumption~\ref{ass2} paves the way for the application of a reparametrization trick, which moves the dependence on $\vct\theta_{l/d}$ from the probability distribution to the sample space of a new random variable and, at the same time, makes this dependence linear. 
This shift allows us to reparametrize the expectations in $\bar c_{l/d}$ as follows:
\begin{equation}
	\overline c_{l/d}(\vct \theta_l,\vct \theta_d)=\mathbb E[c_{l/d}(\mat V_l\vct\theta_l,\mat V_d\vct\theta_d)]\,.
  \label{eq:proof_c}
\end{equation}
Note that Assumption~\ref{ass2} is \emph{not} too restrictive, as many known distributions satisfy the provided conditions (\eg{}, at least all those within the location-scale family such as Gaussian, Laplace, uniform, Cauchy, Weibull, exponential and many others). Indeed, if \eg{} $p_l$ is in the location-scale family, then $\vct w$ is equivalent in distribution to $\vct\mu+\text{diag}(\vct \sigma)\vct z$, where $\text{diag}(\cdot)$ denotes a diagonal matrix with diagonal given by the argument, $\vct\mu\in\mathbb R^{s_l}$ is the \emph{location} parameter, $\vct\sigma\in\mathbb R^{s_d}_{++}$ is the positive, \emph{scale} parameter, and $\vct z$ is a random variable belonging to the same family with standard parametrization (\ie{} location zero and unit scale). 
By setting $\vct\theta_l=(\vct \mu\T,\vct\sigma\T)\T$ and $\mat V_l=[\mat I, \text{diag}(\vct z)]$ we have that
$\mat V_l\vct\theta_l=\vct\mu+\text{diag}(\vct \sigma)\vct z=\vct\mu+\text{diag}(\vct z)\vct \sigma$, which is equivalent in distribution to $\vct w$ as required by Assumption~\ref{ass2}.

In order to be able to rewrite the pseudo-Jacobian $\overline J_{\vct r}$ of our game in terms of $J_{\vct r}$, we further require $c_{l/d}$ to be twice differentiable. This also implies the satisfaction of condition ($i$) in Assumption~\ref{ass1}, \ie{}, the twice differentiability of $\overline c_{l/d}$. 
To satisfy Assumption~\ref{ass1}, as required by Theorem \ref{thm:unique_jr}, we also assume $c_{l/d}$ to be convex. For this to hold, it is sufficient to assume the convexity of regularizers and losses.
This, in conjunction with the linearity of $\mat V_{l/d}\vct\theta_{l/d}$, will then imply (ii-iii) in Assumption~\ref{ass1}.
These conditions are summarized in the following assumption.
\begin{assumption}
\label{ass3}
  For all values of $\vct w$ and $\dot{\mat X}$ sampled from $p_l$ and $p_d$, respectively, the following conditions are satisfied:
  \begin{enumerate}[(i)]
    \item regularizers $\Omega_{l/d}$ are strongly convex and twice differentiable  at $(\vct w,\dot{\mat X})$
    \item for all $y\in\set Y$ and $i\in\{1,\dots,\con n\}$, loss functions $\ell_{l/d}(\cdot,y)$ are convex in $\mathbb R$, and twice differentiable at $\vct w\T\dot{\vct x}_i$.
  \end{enumerate}
\end{assumption}

Under Assumptions~\ref{ass2} and \ref{ass3}, we can finally compute the pseudo-Jacobian $\overline J_{\vct r}$ of our game in terms of the pseudo-Jacobian $J_{\vct r}$ of the underlying prediction game:
\begin{equation}
	\overline J_{\vct r}(\vct \theta_l,\vct \theta_d)=\mathbb E\left[ \mat V\T J_{\vct r}(\mat V_l\vct\theta_l,\mat V_d\vct \theta_d)\mat V \right]\,.
    \label{eq:oJr}
\end{equation}
Here, matrix $\mat V$ is the result of the application of the chain rule for the derivatives in \eqref{eq:Jr}, and it is a block-diagonal \emph{random} matrix defined as
\[
    \mat V=\bmat{\mat V_l&\mat 0\\\mat 0&\mat V_d}\,.
\]

To ensure the positive definiteness of $\overline J_{\vct r}$ we require some additional conditions given in Assumption~\ref{ass4}, which depend on the following quantities:
\begin{align*}
    \lambda_l&=\inf_{\vct\theta_l\in\Theta_l}\lambda_\text{min}(\mathbb E[\mat V_l\T\nabla^2 \Omega_l(\mat V_l\vct\theta_l)\mat V_l])\,,\\
    \lambda_d&=\inf_{\vct\theta_d\in\Theta_d}\lambda_\text{min}(\mathbb E[\mat V_d\T\nabla^2 \Omega_d(\mat V_d\vct\theta_d)\mat V_d])\,,\\
    Q(\vct\theta_l,\vct\theta_d)&=\sum_i\mathbb E[\psi_i(\vct\theta_l\T\mat V_l\T\mat V_d^{(i)}\vct\theta_d)\mat V_d^{(i)}\mat V_l\T]\,,
			\end{align*}
            where $\psi_i(z)=\frac{d}{dz}\ell_l(z,y_i)+\frac{d}{dz}\ell_d(z,y_i)$, $\nabla^2$ is the Hessian operator, $\lambda_\text{min}(\cdot)$ is the smallest eigenvalue of the matrix given as argument, and $\mat V_d^{(i)}$ is the submatrix of $\mat V_d$ corresponding to $\dot{\vct x}_i$, \ie{}
            $\mat V_d^{(i)}\vct\theta_d$ is equivalent in distribution to $\dot{\vct x}_i$.
            Note that $\lambda_{l/d}$ are finite since $\Theta_{l/d}$ are compact spaces.

\begin{assumption}\hfill
  \label{ass4}
  \begin{enumerate}[(i)]
    \item for all $y\in\set Y$, $i\in\{1,\dots,\con n\}$, and for almost all values of $\vct w$ and $\dot{\mat X}$ sampled from $p_l$ and $p_d$, respectively,
			\[
				\ell_l''(\vct w\T\dot{\vct x}_i,y)=\ell_d''(\vct w\T\dot{\vct x}_i,y)\,,
			\]
        \item the players' regularization parameters $\rho_{l/d}$ satisfy
			\[
        \rho_l\rho_d>\frac{\tau}{4\lambda_l\lambda_d}\,,
			\]
        where $\displaystyle
        \tau=\sup_{\vct\theta_{l/d}\in\Theta_{l/d}}\lambda_\text{max}\left( Q(\vct\theta_l,\vct\theta_d) Q(\vct\theta_l,\vct\theta_d)\T\right)\,,
			$
    \end{enumerate}
\end{assumption}
\noindent Here, $\ell_{l/d}''(z,y)=\frac{d^2}{dz^2}\ell_{l/d}(z,y)$ and $\lambda_\text{max}(\cdot)$ returns the largest eigenvalue of the matrix given as argument.

The subsequent lemma states that the positive definiteness of $\overline J_{\vct r}$ is implied by Assumptions \ref{ass2}-\ref{ass4}:
\begin{lemma}
\label{lem:psd}
  If a randomized prediction game satisfies Assumptions \ref{ass2}--\ref{ass4} then
  the pseudo-Jacobian $\overline J_{\vct r}(\vct\theta_l,\vct\theta_d)$ is 
  positive definite for all $(\vct\theta_l,\vct\theta_d)\in\Theta_l\times\Theta_d$ by taking $\vct r=(1,1)\T$.
\end{lemma}
\begin{proof}
    By substituting \eqref{eq:Jr_decomp} into \eqref{eq:oJr} and unfolding the derivatives we can rewirte $\overline J_{\vct r}(\vct\theta_l,\vct\theta_d)$, after simple algebraic manipulations, as the sum of the following matrices
    \[
        \mat J^{(1)}=\mathbb E\left[\sum_i \bmat{\ell''_{l,i}\mat I&\mat 0\\\mat 0&\ell''_{d,i}\mat I}
        \bmat{\mat V_l\T\mat V_d^{(i)}\vct\theta_d\\\mat V_d^{(i)\top}\mat V_l\vct\theta_l}
    \bmat{\mat V_l\T\mat V_d^{(i)}\vct\theta_d\\\mat V_d^{(i)\top}\mat V_l\vct\theta_l}\T\right]
    \]
    \[
        \mat J^{(2)}=\bmat{\rho_l\mathbb E[\mat V_l\T\nabla^2\Omega_l(\mat V_l\vct\theta_l)\mat V_l]&\sum_i\mathbb E[\ell'_{l,i}\mat V_l\T\mat V_d^{(i)}]\\
        \sum_i\mathbb E[\ell'_{d,i}\mat V_d^{(i)\top}\mat V_l]&\rho_d\mathbb E[\mat V_d\T\nabla^2\Omega_d(\mat V_d\vct\theta_d)\mat V_d]
            }
    \]
    where we wrote $\ell''_{l,i}$ for $\ell''_l(\vct\theta_l\T\mat V_l\T\mat V_d^{(i)}\vct\theta_d,y_i)$, and $\ell''_{d,i}$ for $\ell''_d(\vct\theta_l\T\mat V_l\T\mat V_d^{(i)}\vct\theta_d,y_i)$. Similarly, we wrote $\ell'_{l,i}$ and $\ell'_{d,i}$ for the first-order derivatives.

    It is clear from the structure of $\mat J^{(1)}$ that it is positive semidefinite for any $\vct\theta_{l/d}\in\Theta_{l/d}$ if $\ell''_{l,i}=\ell''_{d,i}$ holds almost surely. Therefore, it sufficies to show that $\mat J^{(2)}$ is positive definite to prove that $\overline J_{\vct r}(\vct\theta_l,\vct\theta_d)$ is positive definite. Consider the following matrix
    \[
        \mat H=\bmat{2\rho_l\lambda_l\mat I&Q(\vct\theta_l,\vct\theta_d)\T\\
            Q(\vct\theta_l,\vct\theta_d)&2\rho_d\lambda_d\mat I
            }\,.
    \]
    For any $\vct t=(\vct t_l\T,\vct t_d\T)\T\neq \vct 0$ we have
    \begin{multline*}
    \vct t\T\mat H\vct t=2\rho_l\lambda_l\Vert\vct t_l\Vert^2+2\rho_d\lambda_d\Vert\vct t_d\Vert^2+2\vct t_d\T Q(\vct\theta_l,\vct\theta_d)\vct t_l\\
    \leq \vct t\T(\mat J^{(2)}+\mat J^{(2)\T})\vct t\,,
    \end{multline*}
    where we used the definition of $Q$ and the inequalities 
    \begin{align*}
        \lambda_l\Vert\vct t_l\Vert^2&\leq\vct t_l\T\mathbb E[\mat V_l\T\nabla^2\Omega_l(\mat V_l\vct\theta_l)\mat V_l]\vct t_l\\
        \lambda_d\Vert\vct t_d\Vert^2&\leq\vct t_d\T\mathbb E[\mat V_d\T\nabla^2\Omega_d(\mat V_d\vct\theta_d)\mat V_d]\vct t_d\,,
    \end{align*}
    which follow from the definitions of $\lambda_{l/d}$. Accordingly, we can prove the positive definiteness of $\mat J^{(2)}$ by showing the positive definitness of $\mat H$. To this end, we proceed by showing that all roots of the characteristic polinomial $\det(\mat H-\lambda \mat I)$ of $\mat H$ are positive. By properties of the determinant\footnote{$\det\bmat{a \mat I&\mat B\T\\\mat B&d \mat I}=\det(a\mat I)\det(d\mat I-\frac{1}{a}\mat B\mat B\T)$ and if $\mat U\mat S\mat U\T$ is the eigendecomposition of $\mat B\mat B\T$ then the latter deterimnant becomes $\det(\mat U(d\mat I-\frac{1}{a}\mat S)\mat U\T)=\det(d\mat I-\frac{1}{a}\mat S)$  } we have
    \begin{multline*}
        \det(\mat H-\lambda\mat I)=\det( (2\rho_l\lambda_l-\lambda)\mat I)\\
        \cdot\det\left((2\rho_d\lambda_d-\lambda)\mat I-\frac{\mat S}{2\rho_l\lambda_l-\lambda}\right)\,,
    \end{multline*}
    where $\mat S$ is a diagonal matrix with the eigenvalues of $Q(\vct\theta_l,\vct\theta_d)Q(\vct\theta_l,\vct\theta_d)\T$.
    The roots of the first determinant term are all equal to $2\rho_l\lambda_l$, which is positive because $\rho_l>0$ by construction and $\lambda_l>0$ follows from the strong-convexity of $\Omega_l$ in Assumption~\ref{ass3}-i.
    As for the second determinant term, take the $i$th diagonal element $\mat S_{ii}$ of $\mat S$. Then two roots are given by the solution of the following quadratic polinomial
    \[
        \lambda^2-2\lambda(\rho_l\lambda_l+\rho_d\lambda_d)+4\rho_l\rho_d\lambda_l\lambda_d-\mat S_{ii}=0\,,
    \]
    which are given by
    \[
        \lambda^{(i)}_{1,2}=\rho_l\lambda_l+\rho_d\lambda_d\pm\sqrt{(\rho_l\lambda_l-\rho_d\lambda_d)^2+\mat S_{ii}}\,.
    \]
    Among the two, $\lambda^{(i)}_2$ (the one with the minus) is the smallest one, which is strictly positive if
    $\rho_l\rho_d>\frac{\mat S_{ii}}{4\lambda_l\lambda_d}$.
    Since the condition has to hold for any choice of the eigenvalue $\mat S_{ii}$ in the right-hand-side of the inequality, we take the maximum one $\max_i\mat S_{ii}$, which coincides with $\lambda_\text{max}(Q(\vct\theta_l,\vct\theta_d)Q(\vct\theta_l,\vct\theta_d)\T)$. We further maximize the right-hand-side with respect to $(\vct\theta_l,\vct\theta_d)\in\Theta_l\times\Theta_d$, because we want the result to hold for any parametrization. Therefrom we recover the variable $\tau$ and the condition (ii).
\end{proof}

We finally use this lemma in conjunction to Theorem~\ref{thm:unique_jr} to prove the uniqueness of the Nash equilibrium of a randomized prediction game satisfying Assumptions \ref{ass2}-\ref{ass4}.
\begin{theorem}[Uniqueness]
  \label{thm:uniqueness}
  A randomized prediction game satisfying Assumptions \ref{ass2}--\ref{ass4} has a unique Nash equilibrium.
\end{theorem}
\begin{proof}
    From Assumption~\ref{ass3} it follows that $c_{l/d}$ are twice differentiable and, hence, also $\overline c_{l/d}$ is twice-differentiable and admits the pseudo-Jacobian. Moreover, by Lemma~\ref{lem:psd} the pseudo-Jacobian $\overline J_{\vct r}$ is positive definite.
    It is also easy to see from \eqref{eq:proof_c} that $\bar c_l(\cdot;\vct\theta_d)$ is convex in $\Theta_l$ for all $\vct\theta_d\in\Theta_d$. Indeed, $c_{l}$ is convex in $\vct w$ in view of Assumption \ref{ass3}, and $\mat V_l\vct\theta_l$ is linear in $\vct\theta_l$. Therefore, $c_l(\mat V_l\vct\theta_l,\mat V_d\vct\theta_d)$ is convex with respect to $\vct\theta_l$, and since expectations preserve convexity, it follows  that $\bar c_l(\cdot;\vct\theta_d)$ is convex in $\Theta_l$ as required. By the same arguments, also the convexity of $\bar c_d(\vct\theta_l;\cdot)$ in $\Theta_d$ holds. Hence, Assumption \ref{ass1} holds and Theorem \ref{thm:unique_jr} applies to prove the uniqueness of the Nash equilibrium.
\end{proof}
